\documentclass[10pt,twocolumn,letterpaper]{article}

\usepackage{cvpr}              

\usepackage{amssymb}

\usepackage{xcolor}

\definecolor{avgreen}{HTML}{93C47D}   
\definecolor{avred}{HTML}{EA9999}     

\definecolor{cvprblue}{rgb}{0.21,0.49,0.74}
\usepackage[pagebackref,breaklinks,colorlinks,allcolors=cvprblue]{hyperref}
\usepackage{pifont}
\usepackage{multirow} 
\newcommand{\cmark}{\ding{51}}
\newcommand{\xmark}{\ding{55}}
\usepackage[table]{xcolor}
\definecolor{darkgreen}{RGB}{0,120,0}
\definecolor{darkred}{RGB}{150,0,0}

\usepackage{longtable}
\usepackage{booktabs}
\usepackage{enumitem}

\usepackage{tikz}

\definecolor{sepbar}{HTML}{D98C2F} 
\usepackage{subcaption}

\def\paperID{} 
\def\confName{CVPR}
\def\confYear{2026}

\title{See, Hear, and Understand: Benchmarking Audiovisual Human Speech Understanding in Multimodal Large Language Models}

\author{Le Thien Phuc Nguyen\textsuperscript{1}\footnotemark[1] \quad
    Zhuoran Yu\textsuperscript{1}\thanks{Equal Contribution} \quad
    Samuel Low Yu Hang\textsuperscript{2} \quad
    Subin An\textsuperscript{2} \\
    Jeongkik Lee\textsuperscript{2} \quad
    Yohan Ban\textsuperscript{2} \quad
    SeungEun Chung\textsuperscript{2} \quad
    Thanh-Huy Nguyen\textsuperscript{3} \\
    JuWan Maeng \textsuperscript{2} \quad
    Soochahn Lee\textsuperscript{2} \quad
    Yong Jae Lee\textsuperscript{1}\\[1em]
    {\normalfont
      \textsuperscript{1}University of Wisconsin-Madison\quad
      \textsuperscript{2}Kookmin University \quad
      \textsuperscript{3}Carnegie Mellon University
    }\\
    \normalfont \url{https://plnguyen2908.github.io/AV-SpeakerBench-project-page/}
}

\begin{document}
\maketitle
\begin{abstract}
Multimodal large language models (MLLMs) are expected to jointly interpret vision, audio, and language, yet existing video benchmarks rarely assess fine-grained reasoning about human speech. Many tasks remain visually solvable or only coarsely evaluate speech, offering limited insight into whether models can align who speaks, what is said, and when it occurs. We introduce AV-SpeakerBench, a curated benchmark of 3,212 multiple-choice questions focused on speaker-centric audiovisual reasoning in real-world videos. It features: (1) a speaker-centered formulation that treats speakers—not scenes—as the core reasoning unit; (2) fusion-grounded question design embedding audiovisual dependencies into question semantics; and (3) expert-curated annotations ensuring temporal precision and cross-modal validity. Comprehensive evaluations show that the Gemini family consistently outperforms open-source systems, with Gemini 2.5 Pro achieving the best results. Among open models, Qwen3-Omni-30B approaches Gemini 2.0 Flash but remains far behind Gemini 2.5 Pro, primarily due to weaker audiovisual fusion rather than visual perception. We believe AV-SpeakerBench establishes a rigorous foundation for advancing fine-grained audiovisual reasoning in future multimodal systems.
\end{abstract}
\vspace{-10pt}
\section{Introduction}
\label{sec:intro}
\vspace{-3pt}

Multimodal large language models (MLLMs) have rapidly evolved in recent years, extending language models into image~\cite{liu2023visual, dai2023instructblip, zhu2023minigpt, wang2024qwen2}, video~\cite{lin2024video, damonlpsg2023videollama, wang2024qwen2}, and audio understanding~\cite{kong2024audio, ghosh2024gama, zhang2023speechgpt}. As this evolution continues, recent efforts have moved beyond pairwise modality fusion toward building omni-models that aim to jointly process vision, audio, and language in a unified architecture~\cite{xu2025qwen2, xu2025qwen3, su2023pandagpt, han2024onellm}. Such capability is essential for real-world applications like video dialog agents, meeting transcription systems, and human–AI interaction platforms, where the model must see, hear, and reason over multiple signals simultaneously. However, evaluating whether current models can truly integrate multiple modalities--rather than treat one as the dominant source--remains an open challenge.

\begin{figure}[t]
    \centering
    \includegraphics[width=\linewidth]{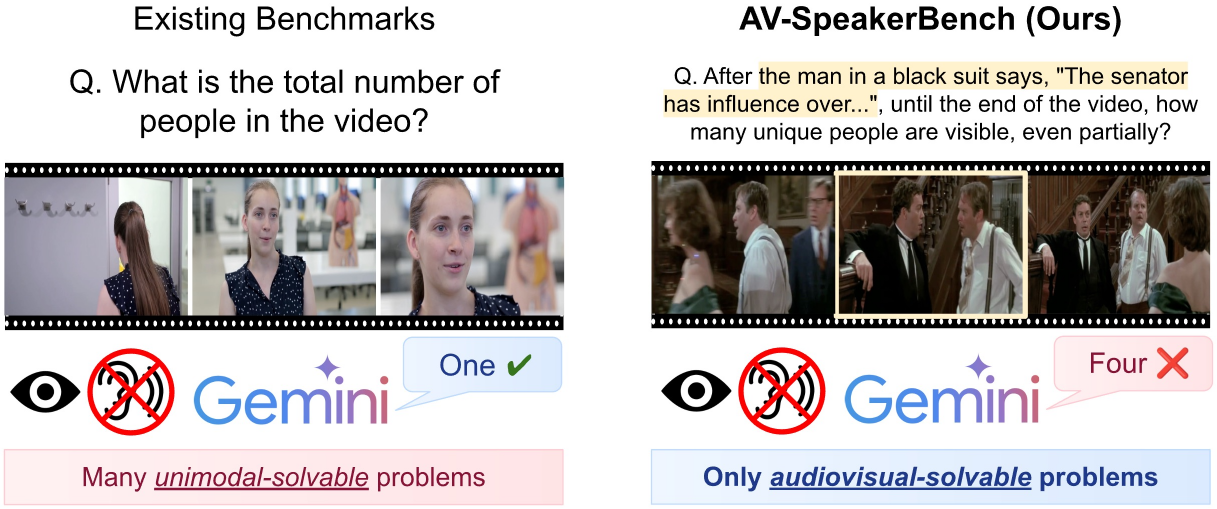}
    \caption{\textbf{Motivation of AV-SpeakerBench.} Many existing video benchmarks contain visually solvable questions, allowing strong multimodal models to answer correctly even without audio (left; e.g., Video-MME~\cite{fu2025video}). By contrast, AV-SpeakerBench (right) requires true audiovisual fusion, where answering depends on who speaks, when they speak, and how speech unfolds over time.}
    \label{fig:teaser}
    \vspace{-20pt}
\end{figure}

In particular, audiovisual speaker perception has been a long-standing research problem~\cite{roth2020ava, jung2024talknce, nguyen2025unitalk, nguyen2025laser, wang2024loconet, kim2021look}, yet modern MLLMs are rarely evaluated on this ability. This gap arises for two reasons. First, existing speaker datasets are built around closed-set labels or frame-level supervision, making them incompatible with open-ended, language-based evaluation. Second, current video QA benchmarks seldom focus on speaker-level reasoning: many questions can be solved using a single modality~\cite{fu2025video, sun2024video}, while others emphasize coarse audio–visual matching or non-speech acoustic events~\cite{yang2022avqa, li2022learning, zverev2025vggsounder}. As a result, current benchmarks do not systematically evaluate whether multimodal models can jointly determine who is speaking, what is being said, and in what visual context.

To address this gap, we introduce \textbf{AV-SpeakerBench}, a new benchmark for evaluating fine-grained audiovisual reasoning centered on human speakers in real-world video. AV-SpeakerBench is explicitly designed to test whether multimodal models can jointly interpret visual, auditory, and linguistic information at the speaker level—a capability not captured by existing datasets. Below, we outline the benchmark’s key design principles.

\vspace{-10pt}
\paragraph{Speaker-centric task formulation.}
AV-SpeakerBench frames every question around the \emph{speaker} as the fundamental reasoning unit, shifting evaluation from scene-level understanding to human-centered audiovisual grounding.
Each video includes multiple visible individuals, enabling identity-based questions that require models to determine \emph{who speaks}, \emph{when}, and \emph{under which visual context}.
By spanning diverse conversational settings and speaker configurations—factors known to increase difficulty in audiovisual perception~\cite{roth2020ava, wang2024loconet, nguyen2025laser, jung2024talknce, nguyen2025unitalk}—the benchmark tests whether models can reliably resolve speaker behavior amid visually complex and varied conversational dynamics.

\vspace{-10pt}
\paragraph{Fusion-driven question design.}
AV-SpeakerBench uses a four-choice MCQ format in which auditory and visual cues are jointly encoded in both questions and answer options.
This design ensures that solving each item requires cross-modal fusion—for example, associating spoken phrases with visible speakers, interpreting speech in relation to visual events, or resolving multiple voices in a shared scene.
Together, these constructions reflect how audiovisual understanding naturally relies on coordinating what is heard with what is seen.

\vspace{-10pt}
\paragraph{High-quality human annotation.}
All clips are manually selected and temporally segmented to isolate moments where speech-driven interaction occurs.
Annotators then identify segments that genuinely require audiovisual reasoning—such as aligning an utterance with the correct visible speaker—and compose MCQs grounded in these spans.
Each clip and question undergoes multi-stage expert review to ensure temporal precision and cross-modal validity.

\vspace{-10pt}
\paragraph{Dataset summary and evaluation scope.}

Our IRB-approved AV-SpeakerBench contains 3,212 curated question–answer pairs across 12 task types, all centered on speakers as the core reasoning unit. These tasks span temporal localization, speaker identification, speech-content reasoning, utterance counting, paralinguistic attribute comparison and so on, collectively evaluating a model’s ability to integrate recognition, grounding, and temporal reasoning across audio and vision rather than depend on static or unimodal cues.

\vspace{-10pt}
\paragraph{Experimental findings and implications.}
Comprehensive evaluation across open-source and proprietary models reveals a clear performance gap between current MLLMs and human accuracy. Among open-source models, only recent omni-directional approaches—such as Qwen3-Omni~\cite{xu2025qwen3}—show substantial progress, with the 30B variant reaching parity with Gemini~2.0~Flash~\cite{comanici2025gemini25pushingfrontier}. However, Gemini~2.5~Pro~\cite{comanici2025gemini25pushingfrontier} remains the strongest overall, outperforming all others across nearly all tasks. Our analysis (Section~\ref{sec:analysis}) shows that this advantage comes primarily from stronger audiovisual fusion: Gemini~2.5~Pro consistently gains 10–20\% from audio inputs, whereas Qwen3-Omni’s improvements are modest or sometimes negative. A systematic error-pattern study further reveals that weaknesses in audio perception and temporal grounding are the dominant sources of failure. Together, these findings show that speaker-centered multimodal reasoning remains a core barrier to human-level audiovisual understanding.
\vspace{-5pt}
\section{Related Work}
\vspace{-3pt}

\begin{table*}[t]
\small
\setlength{\tabcolsep}{6pt}
\centering
\resizebox{\linewidth}{!}{
\begin{tabular}{lcccccccccc}
\toprule
\multirow{2}{*}{\textbf{Benchmarks}} &
\multirow{2}{*}{\textbf{Modalities}} &
\multirow{2}{*}{\textbf{\#Videos}} &
\multirow{2}{*}{\textbf{\#QA}} &
\multirow{2}{*}{\textbf{Anno.}} &
\multicolumn{2}{c}{\textbf{Audiovisual Reasoning}} &
\multicolumn{3}{c}{\textbf{Speaker-centric Reasoning}} \\
\cmidrule(lr){6-7}\cmidrule(lr){8-10}
& & & & &
\shortstack{\textbf{AV} \\ \textbf{Attribution}} & \shortstack{\textbf{AV Temporal} \\ \textbf{Localization}} &
\shortstack{\textbf{Speech–Speaker} \\ \textbf{Attribution}} & \shortstack{\textbf{Speech temporal} \\ \textbf{localization}} &
\shortstack{\textbf{Speech} \\ \textbf{Grounding}}\\
\midrule
MSRVTT-QA~\cite{Xu2017MSRVTTQA}        & V   & 2{,}990 & 72{,}821 & A     & \xmark & \xmark & \xmark & \xmark & \xmark \\
ActivityNet-QA~\cite{Yu2019ActivityNetQA}   & V   &   800  &  8{,}000 & M     & \xmark & \xmark & \xmark & \xmark & \xmark \\
MVBench~\cite{Li2024MVBench}           & V   & 3{,}641 &  4{,}000 & A     & \xmark & \xmark & \xmark & \xmark & \xmark \\
Video-Bench~\cite{ning2023videobenchcomprehensivebenchmarktoolkit} & V   & 5{,}917 & 17{,}036 & A\&M  & \xmark & \xmark & \xmark & \xmark & \xmark \\
EgoSchema~\cite{Mangalam2023EgoSchema} & V   & 5{,}063 &  5{,}063 & A\&M  & \xmark & \xmark & \xmark & \xmark & \xmark \\
Video-MME~\cite{fu2025video}        & V   &   900   &  2{,}700 & M     & \xmark & \xmark & \xmark & \xmark & \xmark \\
MMBench-Video~\cite{Fang2024MMBenchQA} & V   &   609   &  1{,}998 & M     & \xmark & \xmark & \xmark & \xmark & \xmark \\
\midrule
AVQA~\cite{yang2022avqa} & A+V & 57{,}000 & 57{,}335 & M     & \cmark & \xmark & \xmark & \xmark & \xmark \\
OmniBench~\cite{li2025omnibenchfutureuniversalomnilanguage} & A+I & \xmark   & 1{,}142  & M     & \cmark & \xmark & \xmark & \xmark & \xmark \\
AV-Odyssey~\cite{gong2024av} & A+I &   220    & 4{,}555  & M     & \cmark & \xmark & \xmark & \xmark & \xmark \\
WorldSense~\cite{hong2025worldsense} & A+V & 1{,}662  & 3{,}172  & M     & \cmark & \cmark & \xmark & \cmark & \xmark \\
Daily-Omni~\cite{zhou2025dailyomniaudiovisualreasoningtemporal} & A+V &   648    & 1{,}197  & A\&M  & \cmark & \cmark & \xmark & \xmark & \xmark \\
\midrule
\textbf{AV-SpeakerBench (Ours)}                           & A+V & 2{,}051  & 3{,}212  & M     & \cmark & \cmark & \cmark & \cmark & \cmark \\
\bottomrule
\end{tabular}
}
\caption{Comparison of multimodal video QA benchmarks. ``A" means audio, ``V" means video, ``I" means image, ``S" means subtitle. “Anno.” is the annotation type (A: automatic, M: manual).}
\label{tab:benchmarks}
\end{table*}

\paragraph{Multimodal Understanding Benchmarks.}
Multimodal evaluation has expanded from image benchmarks—such as VQA~\cite{antol2015vqa, goyal2017making}, chart/document understanding~\cite{masry2022chartqa, mathew2021docvqa, wang2024charxiv, singh2019towards}, and general capability suites~\cite{fu2025mme, li2023seed, liu2024mmbench}—to video benchmarks that introduce temporal structure and richer scene context.
Different video datasets emphasize different abilities: long-form narrative reasoning ~\cite{Mangalam2023EgoSchema}, temporal ordering~\cite{liu2024tempcompass, cai2024temporalbench}, procedural understanding~\cite{tang2019coin, xiao2021next}, egocentric perception~\cite{damen2018scaling, grauman2022ego4d}, world modeling~\cite{hong2025worldsense}, and broad capability aggregation~\cite{Li2024MVBench, fu2025video}.
While existing video benchmarks broaden multimodal evaluation, they rarely require fine-grained integration between visual signals and human speech, nor do they target speaker-centric reasoning. AV-SpeakerBench fills this gap by providing a dedicated benchmark for speaker-level audiovisual understanding.

\vspace{-10pt}
\paragraph{Audiovisual Understanding Benchmarks.}
Audiovisual speech understanding has been studied through both non-speech and speech-focused datasets. Non-speech benchmarks such as AudioSet~\cite{Jort2017AudioSet} and VGGSound~\cite{chen2020vggsoundlargescaleaudiovisualdataset} target event-level acoustic classification, while speech datasets cover specific facets of communication: VoxCeleb~\cite{Nagrani2017VoxCeleb, Chung2018VoxCeleb2} for speaker identity, LRS/LRS3~\cite{afouras2018lrs3tedlargescaledatasetvisual} for transcription, and AVA-ActiveSpeaker / ASW~\cite{roth2020ava, kim2021look} for frame-level speaking and audibility labels. However, these resources rely on closed-set labels or low-level annotations and thus do not evaluate the open-ended multimodal reasoning abilities expected of modern MLLMs.

Recent work has begun exploring audiovisual reasoning for multimodal models, but mainly through task formulations that differ fundamentally from speaker-centric reasoning.
VGGSounder~\cite{zverev2025vggsounder} reannotates VGGSound~\cite{chen2020vggsoundlargescaleaudiovisualdataset} for multi-label audio event classification; AV-Odyssey~\cite{gong2024av} focuses on audio–visual matching tasks—such as pairing images with audio clips or selecting which video aligns with a given sound—rather than reasoning about spoken content.
WorldSense~\cite{hong2025worldsense} similarly frames audiovisual understanding as matching between visual scenes and acoustic cues (e.g., music style or sound summary), and its tasks span a wide range of audio types without centering human speech; OmniBench pairs static images with narrated audio for temporal reconstruction; and Daily-Omni targets high-level scene understanding rather than modeling speaker identity or spoken utterances. We compare AV-SpeakerBench with other benchmarks in Table~\ref{tab:benchmarks}.

Across these benchmarks, audio is used to characterize scenes, events, or background context, but not to bind speech to visible speakers. AV-SpeakerBench instead adopts a speaker-centric formulation that requires aligning spoken content with the people who produce it.

\vspace{-12pt}
\paragraph{Multimodal Large Language Models.}
Multimodal LLMs have progressed from image–text models such as BLIP-2~\cite{li2023blip}, InstructBLIP~\cite{dai2023instructblip}, and LLaVA~\cite{liu2023visual,liu2024improved} to video-oriented systems like Video-LLaMA~\cite{damonlpsg2023videollama, damonlpsg2024videollama2}, VITA~\cite{fu2024vita}, and PandaGPT~\cite{su2023pandagpt}, which introduce temporal encoding for multi-frame reasoning.
Recent omni-modal models—including Gemini~\cite{team2023gemini}, Qwen-Omni~\cite{xu2025qwen3}, Phi-4 Multimodal~\cite{microsoft2025phi4minitechnicalreportcompact}, StreamOmni~\cite{zhang2025stream}, Unified-IO 2~\cite{lu2023unifiedio2scalingautoregressive}, OLA / OneLLM~\cite{liu2025olapushingfrontiersomnimodal, han2024onellm}, and AnyGPT~\cite{zhan2025anygptunifiedmultimodalllm}—aim to unify audio, video, and language through instruction tuning and shared cross-modal alignment. However, existing evaluations largely emphasize visual QA or coarse audio–visual matching, providing limited insight into whether these models can perform fine-grained, speaker-centric audiovisual reasoning.
AV-SpeakerBench directly targets this gap by assessing whether models can jointly resolve who is speaking, what is said, and when it occurs within natural multi-speaker video.
\section{AV-SpeakerBench}
\label{sec:benchmark}

\begin{figure*}[t]
    \centering
    \includegraphics[width=.97\linewidth]{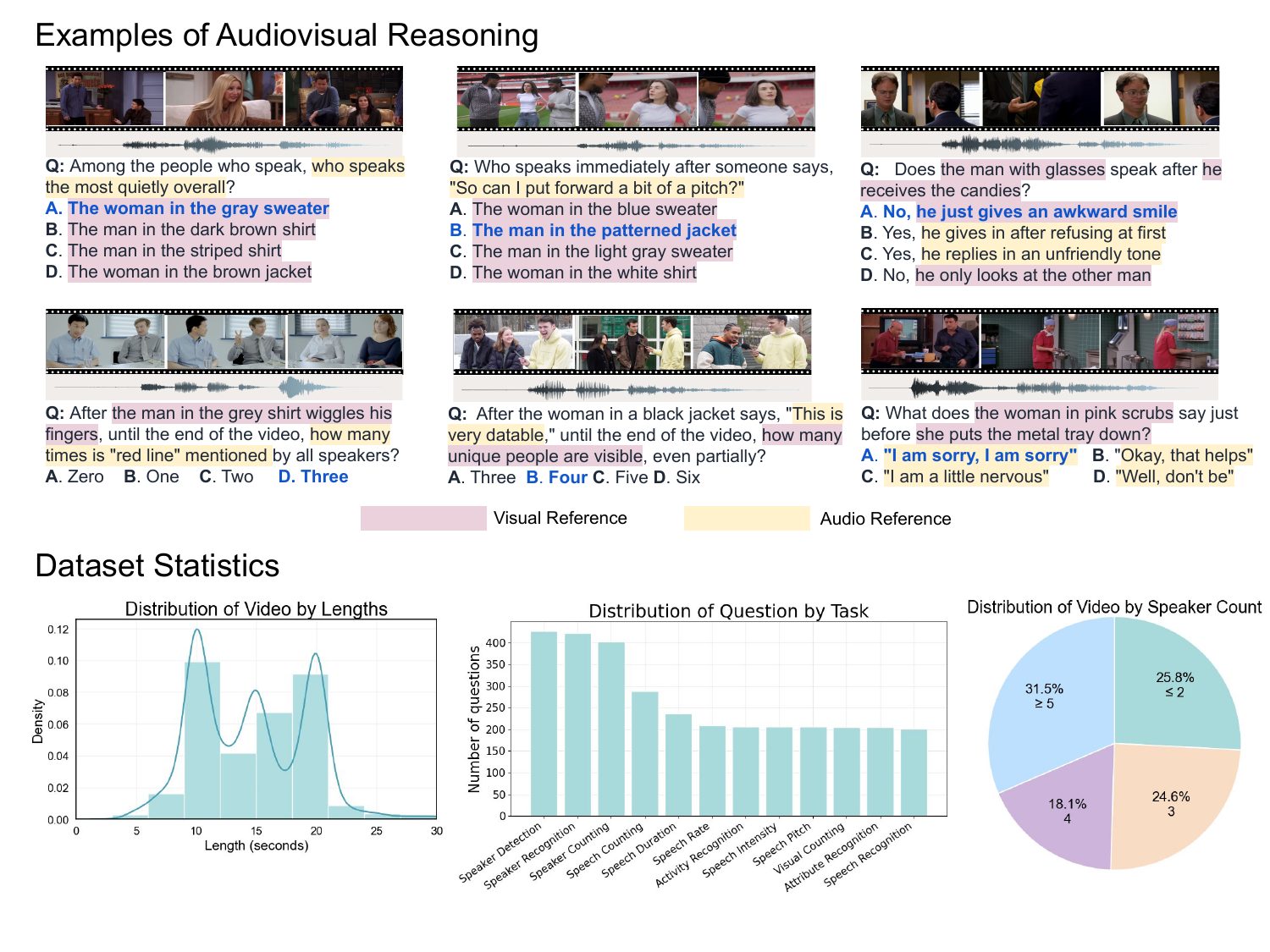}
    \vspace{-1.1cm}
    \caption{
    \textbf{Top: Examples of audiovisual reasoning questions in AV-SpeakerBench.}
    Each question illustrates a distinct way in which audiovisual dependency is enforced—through spoken‐phrase grounding, visual event conditioning, cross-modal temporal localization, or multi-speaker coordination—ensuring that the correct answer cannot be inferred from a single modality.
    \textbf{Bottom: Dataset Distribution.}
    We present the distribution of videos by duration, task category, and visual complexity (measured by the number of unique visible people). Together, these statistics highlight the diversity of conversational scenes and reasoning types represented in \textbf{AV-SpeakerBench}.
    }
    \label{fig:dataset}
\end{figure*}

Recent multimodal video benchmarks such as Video-MME~\cite{fu2025video} have advanced the evaluation of large multimodal models on a wide range of video understanding tasks. However, many existing questions remain largely vision-solvable, such as \textit{``What is the total number of people in the video?''} or \textit{``What clothes is the singer wearing?''}, where the correct answer can be inferred from visual frames alone without attending to the audio stream. Moreover, only a few benchmarks include human speech in their audiovisual evaluation~\cite{yang2022avqa, li2022learning, zverev2025vggsounder}, and those that do remain at a coarse categorical level. For example, VGGSounder~\cite{zverev2025vggsounder} extends VGGSound with multi-label audio–visual classification, where human speech is represented by broad categories such as \textit{male speech}, \textit{female speech}, or \textit{singing}. Such tasks capture the presence of speech but not its linguistic or speaker-specific content, leaving open whether models can reason about \emph{who speaks} and \emph{what is said} in natural audiovisual scenes.

Our goal in constructing AV-SpeakerBench is to fill this gap by focusing on \emph{speaker-centric audiovisual reasoning}. Each question in our benchmark is designed to require joint interpretation of visual and auditory cues in natural human speech scenes. In the following sections, we describe the benchmark’s design principles, data curation and annotation pipeline, and dataset statistics.

\subsection{Fusion-Driven Question Design}
\label{sec:fusion_design}

The central goal of AV-SpeakerBench is to evaluate audiovisual fusion along the temporal dimension—that is, whether a model can coherently align what is seen and heard as events unfold.
All questions adopt a unified multiple-choice (MCQ) format to standardize evaluation and to encode auditory and visual cues compactly within the question and its answer options.
Each item is written so that the correct response requires the model to integrate auditory and visual evidence to reason over temporally ordered actions and utterances.

In our formulation, audiovisual dependency is embedded directly in the textual construction of the question and its options.
The fusion of modalities appears in several complementary ways, including—but not limited to—
(1) linking a spoken phrase to a visible identity (e.g., \textit{When does the man wearing a black T-shirt say ‘What’s going on?’});
(2) cross-conditioning one modality on temporal cues from the other—for example, using visual events to localize speech (e.g, \textit{What does the woman sitting on the porch swing say just before she drinks from her glass?}, or conversely, using auditory events to anchor visual reasoning; \textit{At the moment the man with the black headband says, "We are not cool," how many people are visible?}); and
(3) combining auditory and visual cues in multi-speaker settings (e.g., \textit{After the man in the grey shirt wiggles his fingers, until the end of the video, how many times is ``red line" mentioned by all speakers?}).
Through this formulation, each question enforces temporally grounded multimodal reasoning, ensuring that successful models must align voices, appearances, and contextual events rather than rely on static visual or textual shortcuts. 
Representative examples of different formulation types are illustrated in Figure~\ref{fig:dataset}-top, and a complete set of per-task examples is provided in the Appendix~\ref{supp:example}.

\vspace{-3pt}
\subsection{Dataset Construction}

\label{sec:dataset_construction}

\paragraph{Video sources.}
We collect all video data from YouTube, prioritizing content that naturally features human speech and multi-party interaction.
The majority of clips are sampled from movie clips, as these provide rich audiovisual dynamics such as turn-taking, facial gestures, and speaker transitions within short temporal windows.
To diversify speaking style, visual layout, and recording conditions, we additionally include material from game shows, street interviews, sport interviews, group interviews, podcasts, and vlogs.
All videos are taken from official channels where individuals are clearly and knowingly recorded on camera.

\vspace{-10pt}
\paragraph{Clip selection and task-oriented sampling.}
Unlike datasets where annotators label pre-defined clips, our process begins with segment selection. Annotators watch full videos and identify 5–30s clips that satisfy task requirements and contain meaningful audiovisual variation. Choosing the right moment is crucial: segments with too few speakers (e.g, monologue) or unchanged conversational dynamics make temporal or speaker-grounded questions nearly deterministic (e.g., if only two speakers are visible, \textit{Who speaks after the woman says ‘How are you?’} offers little diagnostic value). Similarly, an anchor condition that does not shift who speaks fails to introduce real temporal grounding. This filtering ensures that all selected clips support genuine temporal localization and cross-modal reasoning rather than superficial pattern matching.

\vspace{-10pt}
\paragraph{Annotation and refinement.}
For each selected clip, annotators compose the question and four answer options following detailed task guidelines specifying the required modality pairings. 
Distractors are drawn from entities, actions, or speech events within the same clip; when insufficient, they are formed by recombining visible or auditory attributes to maintain contextual realism.
All annotators are experienced researchers rather than crowdsourced workers, ensuring close adherence to multimodal reasoning criteria.
Each annotation undergoes a multi-stage refinement pipeline:
(1) an initial review by a separate researcher for clarity and multimodal validity;
(2) linguistic and structural polishing using a language model; and
(3) a final verification round by at least two additional researchers other than the original annotator.
During this process, ambiguous, inconsistent, or trivially solvable examples are filtered out.
This rigorous curation pipeline ensures that all remaining questions exhibit temporal sensitivity, speaker grounding, and robust audiovisual reasoning, providing a balanced and reliable basis for evaluation. Annotation details can be found in Appendix~\ref{supp:annotation}.

\subsection{Dataset Statistics}
\label{sec:dataset_stats}

After refinement, AV-SpeakerBench comprises 3,212 MCQ pairs across 12 task types. All tasks are speaker-centric but probe different aspects of audiovisual understanding, including temporal localization, speaker identification, speech-content retrieval, utterance counting, paralinguistic attribute comparison, and so on. Together, they introduce diverse compositional and temporal challenges that require coordinated multimodal reasoning rather than reliance on static or unimodal cues. Appendix~\ref{supp:example} provides illustrative examples for each task type.

We visualize dataset-wide statistics in Figure~\ref{fig:dataset}-bottom, including the distributions of question types, video durations, and speaker counts.
The dataset spans a wide range of question formulations and task complexities, reflecting the natural diversity of audiovisual reasoning scenarios.
We ensure at least 200 validated examples per task to maintain balanced coverage across reasoning types.
Speaker-related tasks contain more examples, since temporal reasoning about \emph{who speaks and when} can draw on cues from either modality—for instance, determining whether someone speaks after an audio or visual event.
In contrast, tasks centered purely on visual or auditory recognition must rely on the opposite modality to remain genuinely multimodal, which limits the diversity of question formulations rather than their total count.
Although speaker-related tasks therefore include more examples, they also encompass a wider variety of reasoning patterns, and we report per-task accuracy throughout all evaluations to avoid bias toward these categories.

We further analyze the dataset by measuring its \textbf{visual complexity}, quantified by the number of unique visible speakers per clip—a standard indicator of difficulty in human-centric audiovisual benchmarks~\cite{roth2020ava, nguyen2025unitalk, kim2021look}. AV-SpeakerBench covers a wide range of interaction densities, from short dyadic exchanges to multi-party conversations, providing a balanced yet challenging evaluation environment for multimodal large language models.

\subsection{Interpretation and Scope of Evaluation}

Although all question types in AV-SpeakerBench are designed to require audiovisual reasoning in principle, we recognize that strong models may occasionally infer answers from visual cues alone—such as mouth movement or gestures—without explicitly relying on the audio stream. 
We regard this as a natural and desirable behavior rather than a flaw: humans similarly infer speech activity from silent visual context. 
Our benchmark therefore evaluates whether a model can integrate both modalities when available, rather than penalizing instances where one modality alone happens to suffice.
In practice, audio information consistently simplifies reasoning and boosts accuracy, but models achieving correct answers through robust visual understanding remain within the intended evaluation scope.
\vspace{-3pt}
\section{Experiments}
\vspace{-3pt}

\begin{table*}[ht]
\centering
\setlength{\tabcolsep}{6pt}
\renewcommand{\arraystretch}{1.2}
\resizebox{\linewidth}{!}{
\begin{tabular}{lc|ccc|ccc|cccccc|c}
\toprule
\multicolumn{2}{c|}{} &
\multicolumn{3}{c|}{\textbf{Speaker-centric}} &
\multicolumn{3}{c|}{\textbf{Visual-centric}} &
\multicolumn{6}{c|}{\textbf{Audio-centric}} &  \multicolumn{1}{c}{}\\
\midrule
\multicolumn{2}{c|}{\textbf{Task Type}} &
Detection & Recognition & Counting & \shortstack{Attribute\\Recognition}
  & \shortstack{Activity\\Recognition} & Counting &
Recognition & Duration & Pitch & Rate & Intensity & Counting & Overall\\
\midrule
Human Performance & -- & 96.02 & 93.13 & 94.28 & 93.14 & 93.20 & 94.15 & 96.52 & 90.68 & 93.20 & 91.39 & 94.17 & 93.40 & 93.74  \\
\midrule
\multicolumn{14}{c}{\textbf{Proprietary Models}}\\
\midrule
Gemini 2.5 Pro Thinking~\cite{comanici2025gemini25pushingfrontier} & -- & \textbf{81.73} & \textbf{74.15} & \textbf{74.13} & \textbf{72.55} & \textbf{73.30} & \textbf{62.93} & \underline{77.11} & \textbf{78.81} & \textbf{67.48} & \textbf{69.86} & \textbf{71.84} & \textbf{63.89} & \textbf{73.04}\\
Gemini 2.5 Flash Thinking~\cite{comanici2025gemini25pushingfrontier} & -- & \underline{74.71} & \underline{70.62} & \underline{60.95} & \underline{65.59} & \underline{70.39} & \underline{65.85}  & \textbf{78.61} & \underline{69.92} & \underline{66.5} & \underline{67.46} & \underline{65.05} & \underline{58.33} & \underline{67.84}\\
Gemini 2.5 Flash~\cite{comanici2025gemini25pushingfrontier} & -- & 69.79 & 68.24 & 50.75 & 61.76 & 68.45 & 51.22 & 71.64 & 58.47 & 60.68 & 60.29 & 59.71 & 40.97 &  60.27 \\
Gemini 2.5 Flash-Lite~\cite{comanici2025gemini25pushingfrontier} & -- & 45.90 & 52.44 & 39.10 & 48.53 & 51.94 & 43.41 & 55.72 & 51.69 & 51.46 & 46.89 & 54.37 & 36.11 & 47.23 \\
Gemini 2.0 Flash~\cite{comanici2025gemini25pushingfrontier} & -- & 60.19 & 63.51 & 47.51 & 54.9 & 63.59 & 45.85 & 71.14 & 56.78 & 62.62 & 55.98 & 55.83 & 41.67 &  53.21 \\
Gemini 2.0 Flash-Lite~\cite{comanici2025gemini25pushingfrontier} & -- & 56.91 & 57.35 & 38.81 & 49.02 & 55.83 & 45.85 & 67.66 & 52.12 & 52.43 & 51.12 & 56.31 & 33.33 & 51.43 \\
\midrule
\multicolumn{14}{c}{\textbf{Open-sourced Audiovisual Models}}\\
\midrule
Video-LLaMA~\cite{damonlpsg2023videollama} & 7B & 29.51 & 26.07 & \underline{29.35} & \underline{27.23} & \underline{32.52} & 25.37 & \underline{31.84} & \underline{28.81} & \underline{31.55} & 28.23 & 27.67 & 21.53 & 28.21 \\
Video-LLaMA~\cite{damonlpsg2023videollama} & 13B & \underline{30.91} & \underline{29.86} & 25.37 & \underline{27.23} & \textbf{37.38} & 29.76 & 27.86 & 23.31 & 27.67 & \underline{33.49} & \underline{30.58} & \underline{27.08} & \underline{29.11}  \\
Video-LLaMA2~\cite{damonlpsg2024videollama2} & 7B & \textbf{34.19} & \textbf{36.02} & \textbf{31.25} & \textbf{35.29} & \textbf{37.38} &  \textbf{41.46} & 29.85 & \textbf{40.25} & \textbf{49.03} & \textbf{44.02} & \textbf{45.15} & \textbf{31.25} & \textbf{37.67} \\
PandaGPT~\cite{su2023pandagpt} & 7B & 28.34 & 20.61 & 27.36 & 16.18 & 20.87 & \underline{32.20} & \textbf{33.83} & 19.07 & 24.27 & 17.70 & 14.56 & 15.63 & 22.88 \\
PandaGPT~\cite{su2023pandagpt} & 13B & 20.61 & 16.82 & 28.36 & 0.98 & 1.94 & 25.37 & 5.97 & 22.46 & 22.82 & 25.84 & 23.30 & 15.63 & 18.37 \\
\midrule
\multicolumn{14}{c}{\textbf{Open-sourced Omni Models}}\\
\midrule
Unified-IO 2 large~\cite{lu2023unifiedio2scalingautoregressive} & 1B & 24.36 & 27.49 & 23.88 & 24.51 & 23.79 & 22.93 & 21.89 & 22.03 & 25.24 & 31.58 & 31.55 & 34.38 & 26.15 \\
Unified-IO 2 xl~\cite{lu2023unifiedio2scalingautoregressive} & 3B & 28.81 & 28.44 & 31.25 & 30.39 & 26.21 & 23.41 & 24.88 & 30.93 & 22.82 & 26.32 & 30.58 & 31.25 & 27.52  \\
Unified-IO 2 xxl~\cite{lu2023unifiedio2scalingautoregressive} & 7B & 26.70 & 27.25 & 24.63 & 30.39 & 22.82 & 29.76 & 23.28 & 35.59 & 25.24 & 33.49 & 37.38 & 28.8 & 24.97 \\
Phi-4 Multimodal~\cite{microsoft2025phi4minitechnicalreportcompact} & 5.6B & 37.70 & 41.71 & 28.02 & 46.95 & 33.50 & 38.05 & 45.77 & 38.56 & 49.03 & 42.58 & 37.38 & 26.04 & 38.45  \\
OneLLM~\cite{han2024onellm} & 7B & 30.44 & 23.70 & 20.65 & 30.88 & 25.24 & 27.8 & 25.37 & 1.69 & 24.27 & 28.23 & 25.73 & 34.72 & 24.97 \\
AnyGPT~\cite{han2024onellm} & 7B & 24.12 & 4.27 & 24.88 & 0.98 & 19.90 & 19.51 & 15.42 & 0.00 & 0.00 & 0.48 & 2.43 & 22.92 & 12.67\\
VITA~\cite{fu2024vita} & 7B & 32.08 & 34.60 & 29.85 & 37.25 & 36.41 & 28.78 & 32.84 & 37.29 & 38.83 & 35.89 & 37.86 & 28.13 & 33.66 \\
VITA-1.5~\cite{fu2025vita} & 7B  & 32.08 & 36.73 & 32.59 & 38.24 & 35.44 & 35.12 & 32.34 & 45.34 & 43.20 & 44.98 & 36.89 &  29.51 & 36.27\\
Qwen2.5-Omni~\cite{xu2025qwen2} & 3B & 46.37 & \underline{45.26} & \underline{36.57} & 48.04 & 47.57 & \underline{41.95} & 52.74 & \underline{54.24} & 44.17 & 39.23 & 39.23 & 28.12 & 43.52  \\
Qwen2.5-Omni~\cite{xu2025qwen2} & 7B  & \underline{51.76} & 44.31 & \underline{36.57} & \underline{50.49} & \underline{50.49} & \textbf{44.88} & \underline{58.71} & 51.27 & \underline{57.28} & \underline{47.85} & \underline{49.51} & \underline{29.51} & \underline{46.64}  \\
Qwen3-Omni~\cite{xu2025qwen3} & 30B & \textbf{61.83} & \textbf{54.74} & \textbf{46.77} & \textbf{56.86} & \textbf{58.74} & 40.49 & \textbf{68.16} & \textbf{59.32} & \textbf{58.74} & \textbf{55.98} & \textbf{66.02} & \textbf{34.72} & \textbf{54.14} \\
\bottomrule
\end{tabular}
}
\caption{\textbf{Evaluation Results of AV-SpeakerBench}. Our evaluation covers proprietary models, open-sourced audiovisual and omni models. Highest number per model type in each column is \textbf{bolded}, and second-highest number per model type in each column is \underline{underlined}.}
\label{tab:result}
\end{table*}

\subsection{Experiment Setup}

\textbf{Model Selection.} We evaluate multimodal LLMs that natively support audio–video inputs. Among proprietary systems available to us, Gemini~\cite{team2023gemini} is the only model offering true A+V processing; we report both the thinking and non-thinking versions of Gemini~2.5~Flash, and only the thinking version of Gemini~2.5~Pro (the only mode provided). For open-source models, we include all publicly available A+V-capable LLMs across omni-modal and video-first architectures: Qwen2.5-Omni~\cite{xu2025qwen2}, Qwen3-Omni~\cite{xu2025qwen3}, Video-LLaMA~\cite{damonlpsg2023videollama}, Video-LLaMA2~\cite{damonlpsg2024videollama2}, Unified-IO~2~\cite{lu2023unifiedio2scalingautoregressive}, OLA~\cite{liu2025olapushingfrontiersomnimodal}, OneLLM~\cite{han2024onellm}, VITA~\cite{fu2024vita}, VITA-1.5~\cite{fu2025vita}, PandaGPT~\cite{su2023pandagpt}, Phi-4 Multimodal-Instruct~\cite{microsoft2025phi4minitechnicalreportcompact}, and AnyGPT~\cite{zhan2025anygptunifiedmultimodalllm}. For Qwen3-Omni-30B, we report only the non-thinking variant, as the thinking mode takes about 5 minutes per query and would place a full benchmark run beyond our budget. All models are evaluated with their native A+V interfaces using identical inputs and prompts; further details are provided in Appendix~\ref{supp:prompt}.

\begin{figure*}[t]
\centering
\begin{subfigure}[b]{0.6\linewidth}
    \centering
    \includegraphics[height=5cm, width=10cm, keepaspectratio]{pics/modality_ablation.pdf}
    \caption{\textbf{Modality ablation across task types.} 
    Gemini~2.5~Pro demonstrates consistent multimodal gains across most tasks, whereas Qwen3-Omni~30B exhibits limited or even negative audio contributions in certain tasks.}
    \label{fig:modality}
\end{subfigure}%
\hfill
\begin{subfigure}[b]{0.32\linewidth}
    \centering
    \vspace{3mm}
    \includegraphics[height=5cm, keepaspectratio]{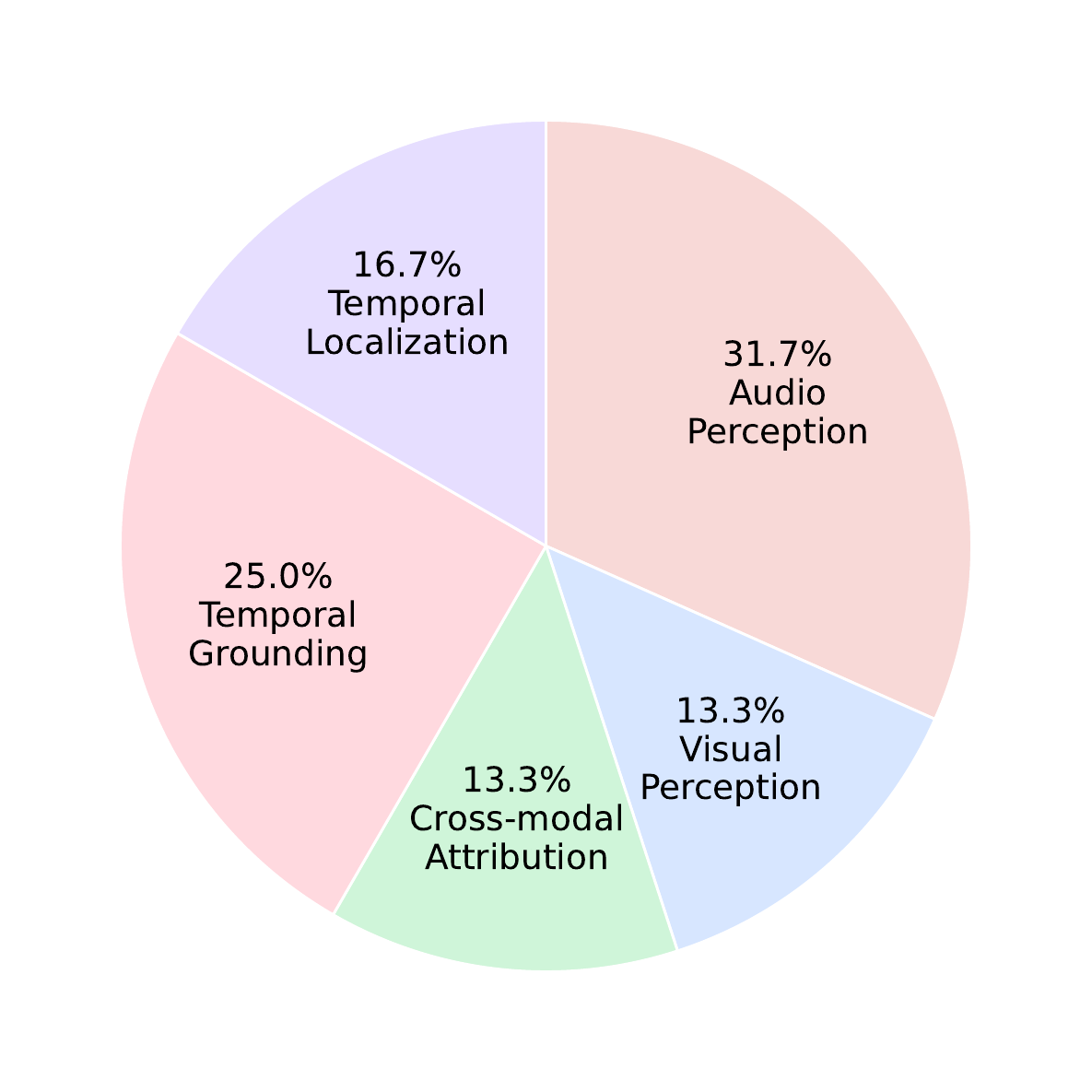}
    \caption{\textbf{Error type distribution across benchmark categories.} Most errors occur in audio perception and temporal reasoning tasks.}
    \label{fig:error}
\end{subfigure}
\vspace{-3mm}
\caption{\textbf{Multimodal ablation and error analysis.}}
\label{fig:ablation_and_error}
\end{figure*}

\noindent\textbf{Frame Sampling.} We follow each model’s default temporal sampling policy. Concretely, we sample 8 frames for Video-LLaMA~\cite{damonlpsg2023videollama}, Video-LLaMA2~\cite{damonlpsg2024videollama2}, Phi-4 Multimodal-Instruct~\cite{microsoft2025phi4minitechnicalreportcompact}, Unified-IO~2~\cite{lu2023unifiedio2scalingautoregressive}, and AnyGPT~\cite{zhan2025anygptunifiedmultimodalllm}; 5 frames for PandaGPT~\cite{su2023pandagpt} and OneLLM~\cite{han2024onellm}; 1 fps capped at 100 frames for VITA~\cite{fu2024vita} and VITA-1.5~\cite{fu2025vita}; and 1 fps for Gemini~\cite{team2023gemini}, Qwen2.5-Omni~\cite{xu2025qwen2}, and Qwen3-Omni~\cite{xu2025qwen3}. Unless otherwise noted, frames are sampled uniformly within each clip. We ablate the performance of these models using different sampling strategies in Appendix~\ref{supp:frame_sampling}.

\subsection{Results}
The full evaluation results are presented in Table~\ref{tab:result}; all results are reported in MCQ accuracy. Here, we summarize the main takeaways.

\noindent\textbf{Existing models remain far from human performance.}
Human evaluation reaches an overall accuracy of 93.74\%, confirming that the questions are clear and naturally solvable through audiovisual reasoning.
In comparison, the strongest model—Gemini 2.5 Pro—achieves only 73.04\%, leaving a gap of over 20 percentage points.
This indicates that current multimodal LLMs still have substantial room for improvement on fine-grained, temporally grounded speaker-centric reasoning.

\noindent\textbf{Gemini models demonstrate stronger performance than open-source models.}
Gemini 2.5 Pro (thinking) delivers the best results on AV-SpeakerBench, outperforming all other models on 11 of the 12 tasks and achieving an overall accuracy of 73.04\%.
Gemini 2.5 Flash (thinking) improves upon the non-thinking Flash variant by 7.57\%, yet still falls notably short of 2.5 Pro.
Since both variants employ the thinking process, this performance gap primarily reflects differences in the underlying model strength, with possible additional gains from how the thinking mechanism interacts with Pro’s larger capacity.

\noindent\textbf{Only recent Omni-series models achieve non-trivial performance on AV-SpeakerBench.}
Earlier open-source multimodal models—such as Video-LLaMA2~\cite{damonlpsg2024videollama2}, PandaGPT~\cite{su2023pandagpt}, OneLLM~\cite{han2024onellm}, Unified-IO~\cite{lu2023unifiedio2scalingautoregressive}, and AnyGPT~\cite{zhan2025anygptunifiedmultimodalllm}—perform poorly on AV-SpeakerBench, often approaching random-guessing accuracy despite supporting both audio and video inputs.
Recent Omni-series models, including Phi-4-Multimodal~\cite{microsoft2025phi4minitechnicalreportcompact}, VITA/VITA-1.5~\cite{fu2024vita, fu2025vita}, Qwen2.5-Omni~\cite{xu2025qwen2}, Qwen3-Omni~\cite{xu2025qwen3}, and OLa~\cite{liu2025olapushingfrontiersomnimodal}, demonstrate substantial improvements, marking the first generation of open-source models capable of handling fine-grained audiovisual reasoning.
Among them, Qwen3-Omni-30B is the strongest, exceeding all other open-source models and even slightly surpassing Gemini 2.0 Flash.
However, it still remains far behind the Gemini 2.5 family, underscoring the difficulty of speaker-centric audiovisual reasoning even for the most advanced open-source LLMs.

\subsection{Analysis}
\label{sec:analysis}

\paragraph{Performance gaps reflect differences in audiovisual fusion capability.}
From Table~\ref{tab:result}, we observe that Gemini 2.5 Pro consistently outperforms Qwen3-Omni-30B across all categories on AV-SpeakerBench.
We investigate this difference by comparing each model’s performance under \emph{vision-only} and \emph{audiovisual} input settings (Figure~\ref{fig:modality}).
Gemini 2.5 Pro exhibits consistent gains of roughly 10–20 percentage points across all tasks when both modalities are available, indicating stable and effective fusion.
In contrast, Qwen3-Omni-30B achieves much smaller gains—and in some tasks, even negative differences—suggesting that audio input does not reliably improve its reasoning.
These results indicate that Gemini’s advantage primarily arises from stronger temporal alignment and cross-modal integration, whereas future progress in speaker-centric audiovisual reasoning will depend on improving fusion robustness rather than merely scaling unimodal perception.

\vspace{-10pt}
\paragraph{Why advanced models may answer some audiovisual questions using only vision.}
Although AV-SpeakerBench is designed to require audiovisual reasoning, we observe that advanced models can occasionally answer certain questions using only visual cues. Human speech naturally produces observable signals—such as mouth motion, gaze shifts, and conversational gestures—that can provide partial evidence about \emph{who is speaking} or \emph{when speech occurs}. As a result, strong models may exploit these cues even in the absence of audio. More information can be found in Appendix~\ref{supp:vision_only}.

Figure~\ref{fig:qualitative}a shows an example where Gemini 2.5 Pro correctly identifies the most active speaker under vision-only input by tracking sustained mouth movement and gesturing patterns. However, these cues are not always reliable. In Figure~\ref{fig:qualitative}b, the same vision-only setting leads the model to a wrong answer because the visible gestures mislead it about the speaker’s rate of speech. When audio is provided (Figure~\ref{fig:qualitative}c), the model correctly resolves the ambiguity, demonstrating that the task genuinely requires multimodal fusion.

Importantly, these visual cues reflect natural properties of human communication rather than dataset bias. \textbf{AV-SpeakerBench does not explicitly penalize unimodal solutions—if a model can reliably infer speech activity from vision alone, this reflects genuine capability (e.g., implicit lip-reading or motion-level reasoning).}
However, as Figures~\ref{fig:qualitative}b–c illustrate, \emph{fusion offers a clear advantage}: audio resolves ambiguities that visual cues alone cannot, and robust performance across the benchmark consistently requires integrating both modalities. Our goal is therefore not to suppress unimodal cues, but to design tasks where multimodal fusion provides the most reliable and general path to correct reasoning. 

\begin{figure*}[t]
    \centering
    \includegraphics[width=0.8\linewidth]{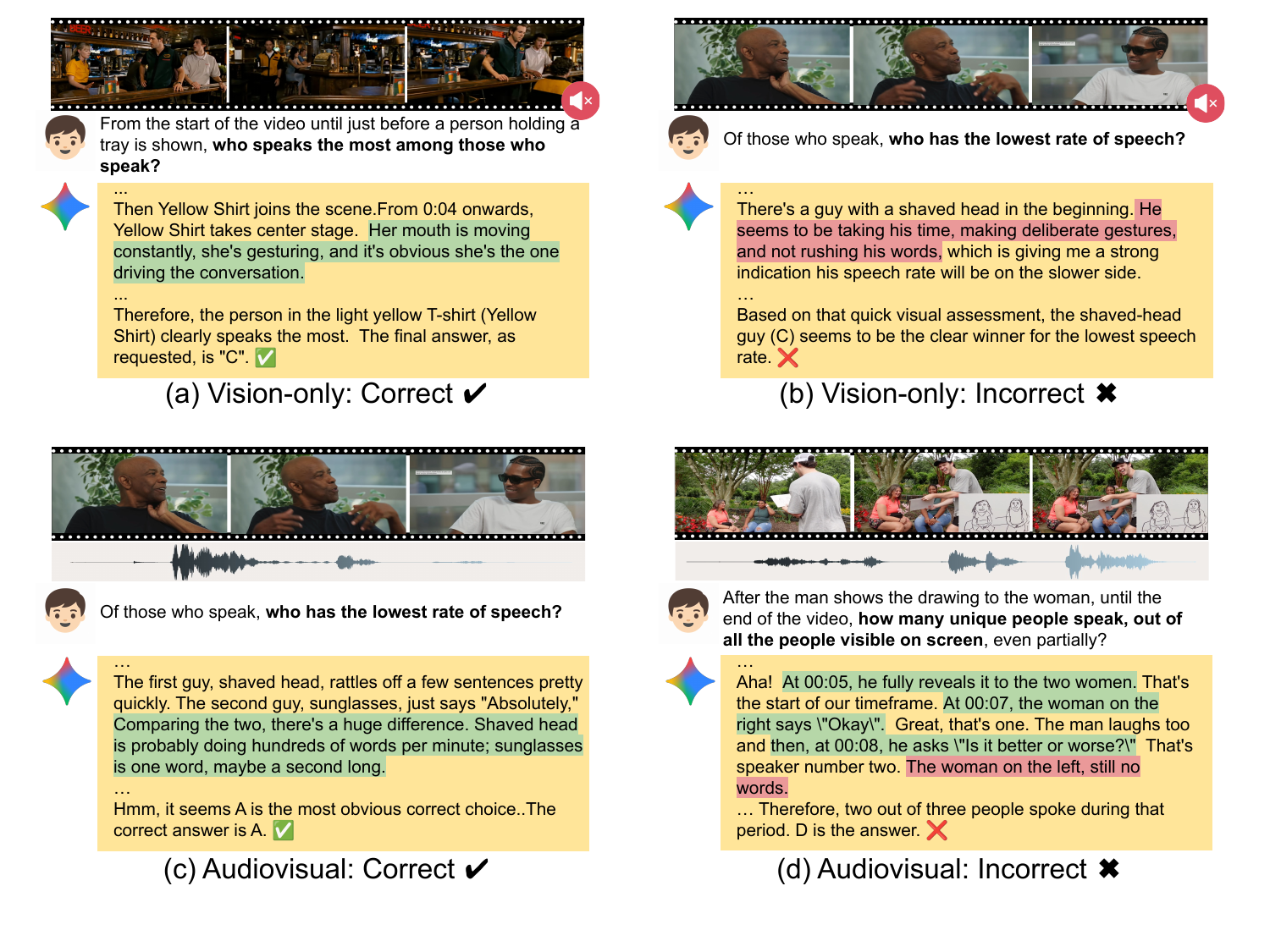}
    \vspace{-20pt}
    \caption{\textbf{Qualitative examples of Gemini 2.5 Pro} reasoning traces on \textbf{AV-SpeakerBench}. \textcolor{darkgreen}{Green} and \textcolor{darkred}{red} highlight colors indicate the model’s correct and incorrect reasoning, respectively. \textbf{(a) Vision-only example answered correctly:} the model identifies the correct speaker by tracking the duration and consistency of mouth movement and conversational gestures, which serve as natural visual cues for inferring who is speaking. \textbf{(b) Vision-only example answered incorrectly:} the model incorrectly associates slower gestures with slower speech, leading to a wrong prediction. \textbf{(c) The same example as (b) but with audio input:} the model correctly identifies the faster speaker once speech-rate evidence becomes available, confirming that the question requires true audiovisual fusion. \textbf{(d) Vision + audio example answered incorrectly:} the model predicts that only one woman speaks while both women say ``Okay'' after the event. Eventually, all three speakers talk after the event, showing residual difficulty in temporal alignment and speaker disambiguation.}

    \label{fig:qualitative}
\end{figure*}

\begin{table}[t]
    \centering
    \resizebox{\linewidth}{!}{
    \begin{tabular}{l|cccc}
    \toprule
        \textbf{Number of Unique People} &  $\le$2 & 3 & 4 & $\ge$ 5\\
    \midrule
        Gemini 2.5 Pro Thinking~\cite{comanici2025gemini25pushingfrontier} & 74.8 & 74.1 & 74.1 & 70.9\\ 
        Gemini 2.5 Flash Thinking~\cite{comanici2025gemini25pushingfrontier} & 71.8 & 68.4 & 66.8 & 65.1 \\ 
        Qwen2.5-Omni 7B~\cite{xu2025qwen2} & 52.0 & 47.8 & 45.3 & 42.7 \\
        Qwen3-Omni 30B~\cite{xu2025qwen3} & 58.3 & 52.9 & 52.0 & 54.4\\
    \bottomrule
    \end{tabular}
    }
    \vspace{-3mm}
    \caption{\textbf{Accuracy by the number of visible people}. Models generally decrease in accuracy as visual complexity increases.}
    \label{tab:acc_by_people}
\end{table}

\vspace{-10pt}
\paragraph{Failure cases for Gemini 2.5 Pro.}
To better understand the remaining challenges for Gemini 2.5 Pro on AV-SpeakerBench, we qualitatively examined five failure cases per task and categorized them into four major types:
\textbf{(1) Visual/Audio perception error}—misperceiving information within a single modality, such as misidentifying a gesture or mishearing a spoken phrase;
\textbf{(2) Cross-modal attribution error}—recognizing unimodal content correctly but mismatching it across modalities, e.g., attributing an utterance to the wrong speaker;
\textbf{(3) Temporal grounding error}—identifying the correct event but applying an incorrect temporal relation, such as reasoning over the \emph{after} segment when the question asks for \emph{before}; and
\textbf{(4) Temporal localization error}—selecting the wrong segment in audio or video, leading to misaligned reasoning.
As summarized in Figure~\ref{fig:error}, most failures arise from \textbf{audio perception} and \textbf{temporal grounding}, indicating that even advanced models struggle to parse overlapping speech and accurately anchor reasoning in time.
Figure~\ref{fig:qualitative}d further illustrates an \emph{audio perception error} where the model misses one of two speakers uttering the same word in close succession. Additional examples for each error type are provided in Appendix~\ref{supp:error_example}.

\vspace{-10pt}
\paragraph{Evaluation by visual complexity.}
To examine how visual complexity impacts model performance, we regroup questions by the number of visible faces, as shown in Table~\ref{tab:acc_by_people}. All models show lower accuracy as the number of people increases, indicating that multi-person scenes create substantial challenges for speaker identification, temporal association, and audiovisual grounding.

\vspace{-7pt}
\section{Conclusion}
\vspace{-3pt}

We introduced AV-SpeakerBench, a benchmark for evaluating fine-grained, speaker-centric audiovisual reasoning in real-world video. Unlike prior multimodal datasets that focus on scene- or event-level understanding, AV-SpeakerBench centers on human speakers and explicitly requires integrating auditory and visual cues. Our experiments show that while Gemini 2.5 Pro exhibits strong fusion and temporal grounding, other models still face clear limitations in multimodal integration, underscoring the need for more robust fusion mechanisms in future multimodal systems. We hope this benchmark provides a foundation for advancing multimodal models toward genuine audiovisual understanding—where language, vision, and speech are jointly reasoned within the same conversational context.

\section{Acknowledgement}
This work was supported in part by NSF IIS2404180, IBM, and Institute of Information \& communications Technology Planning\& Evaluation (IITP) grants funded by the Korea government (MSIT) (No. 2022-0-00871, Development of AI Autonomy and Knowledge Enhancement for AI Agent Collaboration), (No. RS-2022-00187238, Development of Large Korean Language Model Technology for Efficient Pre-training), (No. RS-2025-2543949. Environment-Aware and Domain-Adaptive Multimodal Embodied AI for Real-World Interaction), and (No.RS-2025-02219317, AI Star Fellowship (Kookmin University)).
{
    \small
    \bibliographystyle{ieeenat_fullname}
    \bibliography{main}
}

\clearpage
\setcounter{page}{1}
\maketitlesupplementary
\appendix

\section{Annotation Details}
\label{supp:annotation}
This section outlines the details of our annotation process. 
We first describe the interface used to collect audio–visual annotations, followed by the task guidelines provided to annotators to ensure consistency across examples. 
We then outline the criteria used to select experienced annotators for this work. 
Finally, we include a trivially solvable case to illustrate the intended task clarity and establish a lower bound on annotation difficulty.

\subsection{Annotation Interface}
\begin{figure*}[t]
    \centering
    \includegraphics[width=\linewidth]{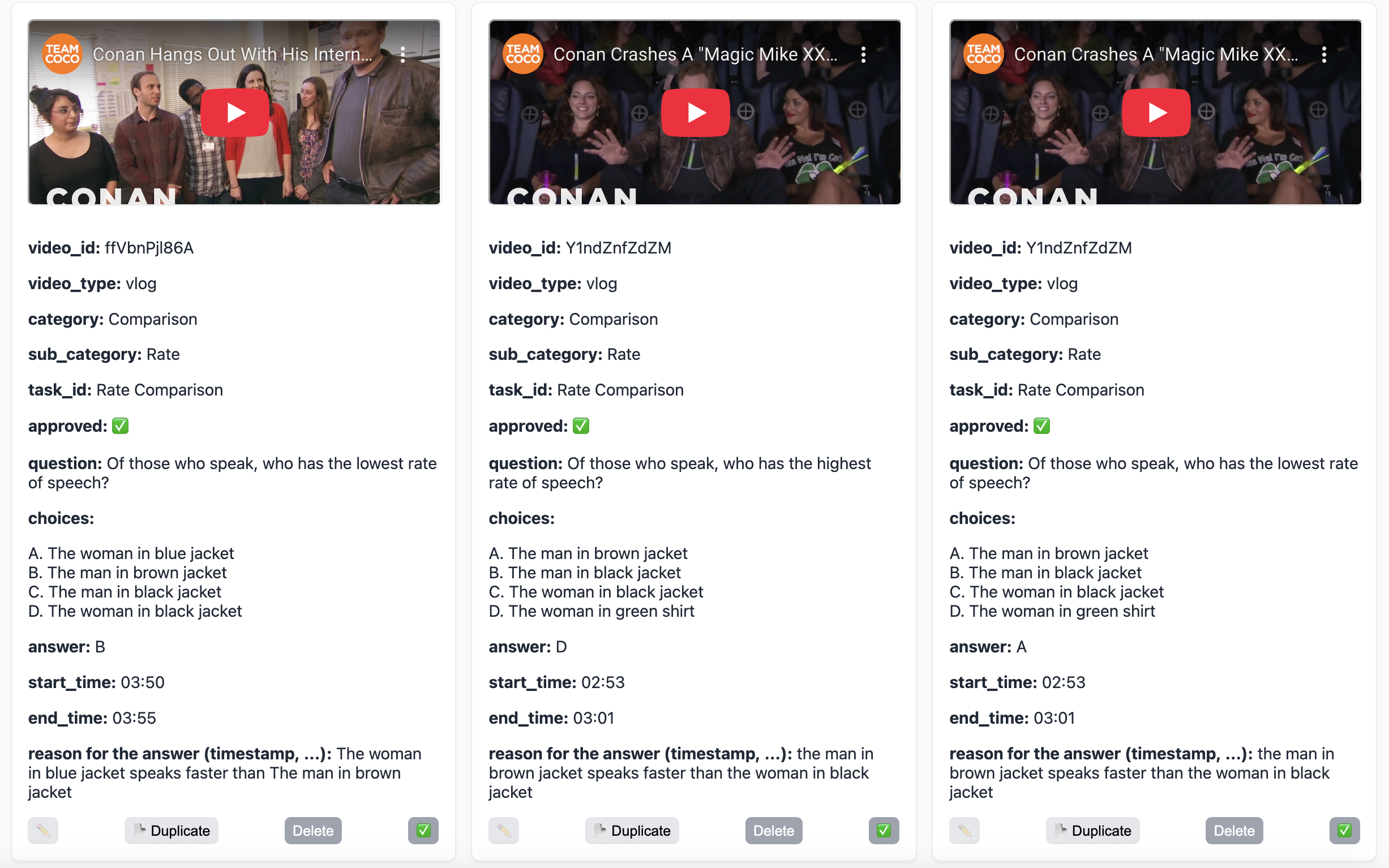}
    \caption{\textbf{Annotation interface for rate–comparison tasks.}
The interface presents annotators with the video clip, metadata (video ID, category, task type), the question, all answer choices, and the selected response. 
Annotators also specify the temporal window used for judgment and provide a brief justification. 
The examples shown correspond to (left) lowest rate of speech, (middle) highest rate of speech, and (right) lowest rate of speech for a different time span within the same video. 
These examples illustrate how annotators validate temporal reasoning by explicitly grounding answers in the video timeline.}
    \label{fig:annotation_interface}
\end{figure*}

Figure~\ref{fig:annotation_interface} shows the web-based interface used to collect multiple-choice audio--visual annotations.
Each example is presented as a self-contained card with a video player at the top and all annotation fields displayed below. 
The card includes immutable metadata (e.g., \texttt{video\_id}, \texttt{video\_type}, \texttt{category}, \texttt{sub\_category}, \texttt{task\_id}) and an \texttt{approved} flag, followed by the natural-language question and four answer options (A--D).
Annotators watch the clip, specify the temporal span used for reasoning via the \texttt{start\_time} and \texttt{end\_time} fields, select the correct answer, and provide a short explanation in the ``reason for the answer'' field, typically referencing visual or auditory evidence.
The researchers responsible for review will examine the reason together with the question and answer choices to determine whether a question should be refined or approved.

\subsection{Annotation Guideline}
\label{supp:taxnonomy}

\subsubsection{General Guidelines}

We provide annotators with a set of general guidelines to ensure consistent and high–quality audio--visual question creation.
Annotators work from a pool of English-language YouTube videos and follow a standardized workflow: each clip is added to the interface, a 5-to-30-second segment is selected, and a single multiple-choice question is authored according to the assigned task type. 
Annotators record the answer, and supply a brief justification.

\paragraph{Video Selection and Clip Duration.}
We provide the following constraints to maintain data quality:

\begin{itemize}
    \item \textbf{Captions.} Select videos without persistent on-screen subtitles or captions that would trivially reveal the spoken content.
    \item \textbf{Content safety.} Exclude clips containing extreme or explicit violence or gore. Mild or non-gratuitous physical conflict is allowed, but highly graphic or disturbing content should be avoided.
    \item \textbf{Language.} All selected videos must be in English.
    \item \textbf{Identity visibility.} Ensure that the queried identity is clearly visible in the foreground and is visually distinguishable from other people in the scene.
    \item \textbf{Discriminative attributes.} When designing questions about a specific person, choose attributes (e.g., clothing, position, actions) that clearly distinguish that person from others in the clip.
    \item \textbf{Contextual distractors.} Construct all answer options from entities, actions, or speech events within the same clip. 
    When there are not enough distinct candidates, form distractors by recombining visible or auditory attributes (e.g., clothes, positions, or phrases) so that each option remains contextually plausible and balanced.
    \item \textbf{Clip length.} Localize each question to a contiguous 5--30\,s segment, depending on the task’s requirements, trimming longer spans to the minimal window that still supports correct reasoning.
\end{itemize}

\subsubsection{Task-specific Guideline}

We provide task-specific requirements in Table~\ref{tab:task_requirements}. 
Annotators apply these templates when constructing multiple-choice questions.
The annotators will add a few questions for us to provide feedback before they scale up the data annotation.

\subsection{Annotator Selection}

To ensure high-quality labels for tasks requiring fine-grained audio--visual reasoning, we restrict annotation to experienced researchers rather than crowd workers. Our annotator team has prior research experience in video understanding, speech processing, or multimodal large language models, enabling them to identify and select clips that are meaningful and informative for the tasks. This careful selection process is crucial for producing labels of sufficient quality and reliability, which would be difficult to achieve with crowd-sourced annotators.

In addition, researchers are trained on the full task taxonomy (see Section~\ref{supp:taxnonomy}) before annotation. 
This ensures that annotators understand the distinction between trivially solvable cases and those requiring joint audio--visual inference, and could apply task definitions precisely. 

\subsection{Eliminating Trivially Solvable Questions}
\begin{figure}[t]
    \centering
    \includegraphics[width=\linewidth]{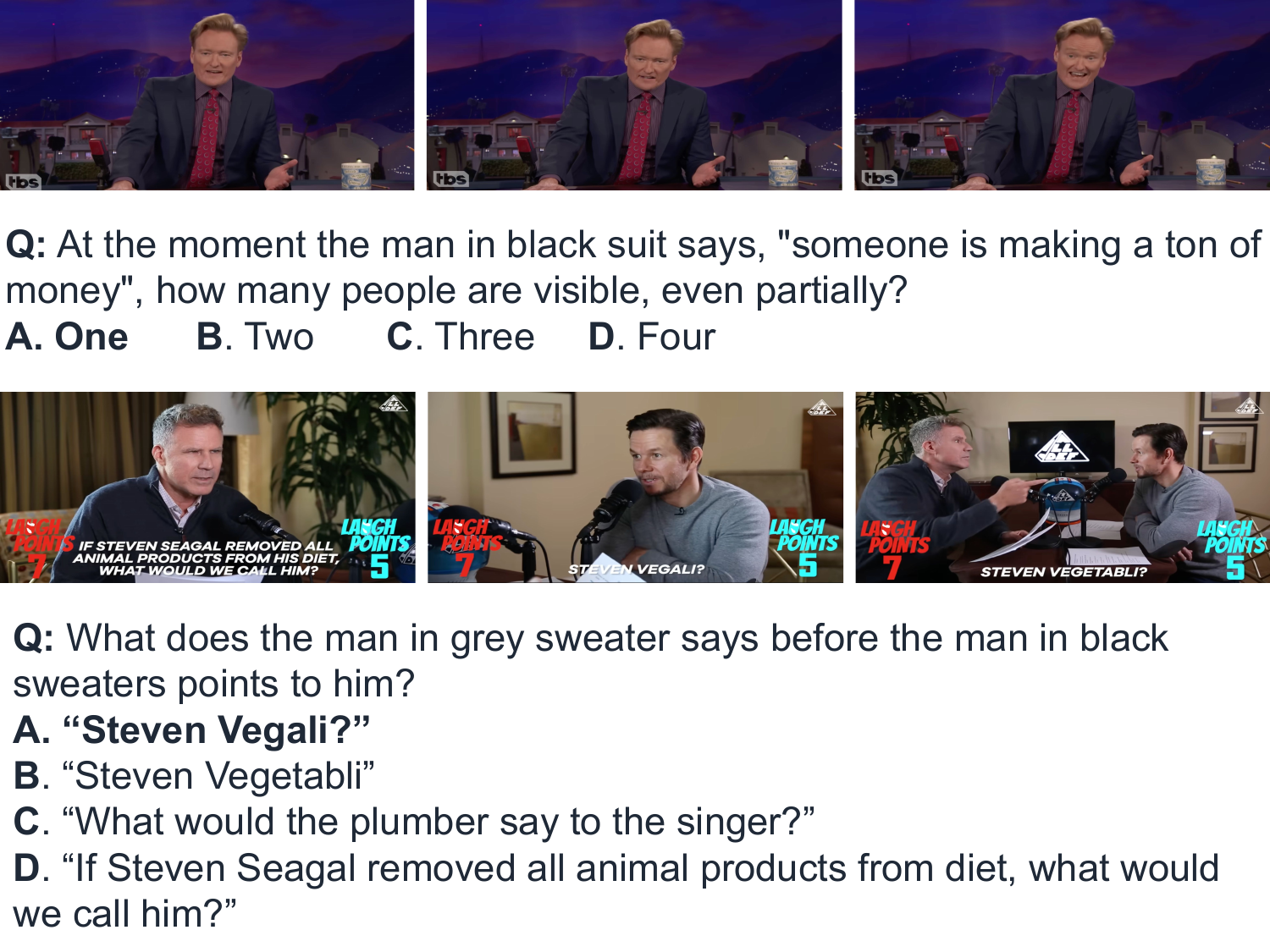}
    \caption{\textbf{Examples of trivially solvable questions removed during filtering.}
    (\textbf{Top}) A moment-specific visibility question becomes trivial because only one person is visible throughout the entire clip, making the answer recoverable without grounding to the referenced utterance. 
    (\textbf{Bottom}) A speech-content question becomes trivial because the spoken line appears as burned-in captions, allowing the answer to be selected without performing audio-based reasoning.}
    \label{fig:trivial_examples_2}
\end{figure}

A key step in our quality-control pipeline is identifying and removing \emph{trivially solvable} questions—items whose answers can be inferred without reference to specific temporal or spatial cues in the clip. These questions arise when the structure of the video or the phrasing of the question makes the correct answer globally obvious, regardless of any localized event. Typical examples include questions where the target event occurs throughout the video or is otherwise easily predictable, providing little meaningful challenge for the fine-grained audio--visual reasoning our tasks are designed to assess.

\begin{itemize}
    \item \textbf{Persistent speaker patterns}: e.g., one person speaks continuously throughout the clip, making questions such as ``Does the man speak when event $X$ happens?'' solvable without checking the requested moment.
    \item \textbf{Constant scene composition}: e.g., the number of visible people remains unchanged across the entire segment, allowing ``How many people are visible when event $X$ occurs?'' to be answered without performing moment-specific counting.
    \item \textbf{Globally obvious answers}: cases where the question refers to a specific moment or segment, but the answer can be determined from a property that holds across the entire clip rather than from the localized event.
    \item \textbf{Always-on video captions}: some videos contain burned-in subtitles throughout the entire clip. 
    In such cases, questions involving spoken content (e.g., ``What does the man say after event $X$?'') become answerable by simply reading the captions rather than performing speech recognition or audio--visual alignment.
    Therefore, these questions fail to evaluate the intended modality.
\end{itemize}

Figure~\ref{fig:trivial_examples_2} shows two representative examples. 
In the first, only a single person is visible across the entire segment, so the answer to a moment-specific visibility question is globally obvious. 
In the second, the spoken line appears verbatim as burned-in captions on the screen, allowing the correct answer to be selected without listening to or timing the utterance. 
Such items do not evaluate the intended audio--visual capabilities and are therefore removed.

\section{Detailed Evaluation}
\label{supp:prompt}

This section provides additional details on our evaluation protocol beyond what
is reported in the main paper. 
We first specify the exact prompts and answer format used when querying multimodal language models on our multiple–choice audio–visual questions (Sec~\ref{supp:eval_prompt}). 
We then describe our procedure for measuring human performance, including annotator recruitment, instructions, and aggregation of responses (Sec~\ref{supp:human_perf}), which serves as a reference ceiling for model accuracy on our benchmark.

\subsection{Evaluation Prompt}
\label{supp:eval_prompt}

\begin{figure}[t]
    \centering
    \includegraphics[width=\linewidth]{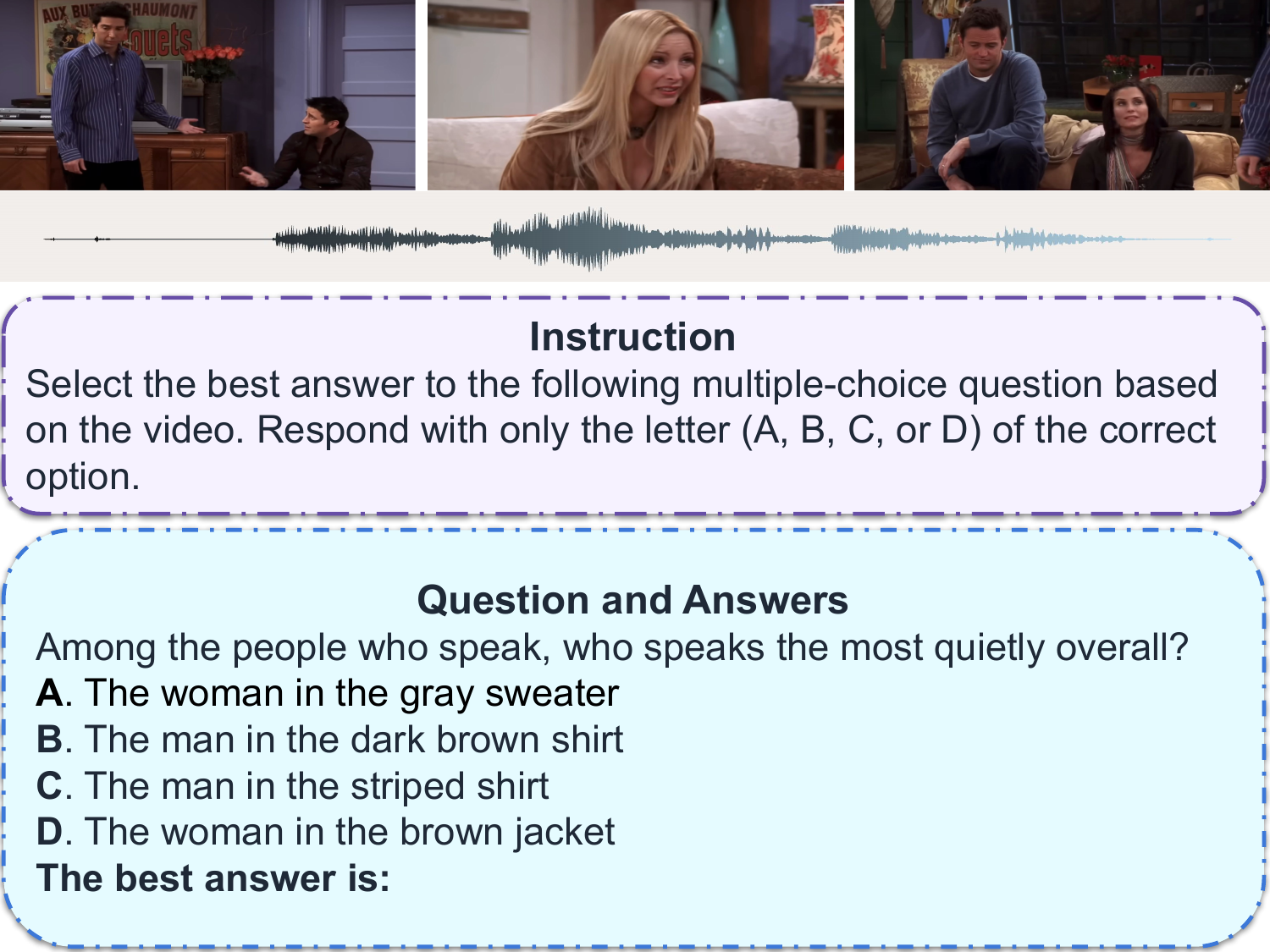}
    \caption{Evaluation prompt used for multimodal LLMs. 
    For each example we show a short video (represented here by keyframes and the waveform), a fixed natural-language instruction, and a multiple-choice question with four options (A–D). 
    Models must answer by outputting only the letter of the correct option.}
    \label{fig:eval_prompt}
\end{figure}

At test time, we query models with a single unified prompt for all tasks.
Figure~\ref{fig:eval_prompt} illustrates the format. 
Each instance consists of (i) the video input (uniformly sampled frames plus the full audio track), (ii) a fixed instruction, and (iii) a multiple-choice question with four options labeled A–D.

The instruction reads:

\medskip
\noindent\textit{``Select the best answer to the following multiple-choice question based on the video. Respond with only the letter (A, B, C, or D) of the correct option.''}
\medskip

The question and answer block then specifies the task (e.g., speech intensity, speaker counting) and lists four candidate answers in natural language. 
We parse the model’s prediction by mapping it to the corresponding option among $\texttt{A}$, $\texttt{B}$, $\texttt{C}$, or $\texttt{D}$. Any response that does not contain a valid option is counted as incorrect. 
This strict, letter-only format helps create deterministic results through standardized instructions across all sub-benchmarks, following prior work~\cite{fu2025video}.

\begin{figure}[t]
    \centering
    \includegraphics[width=\linewidth]{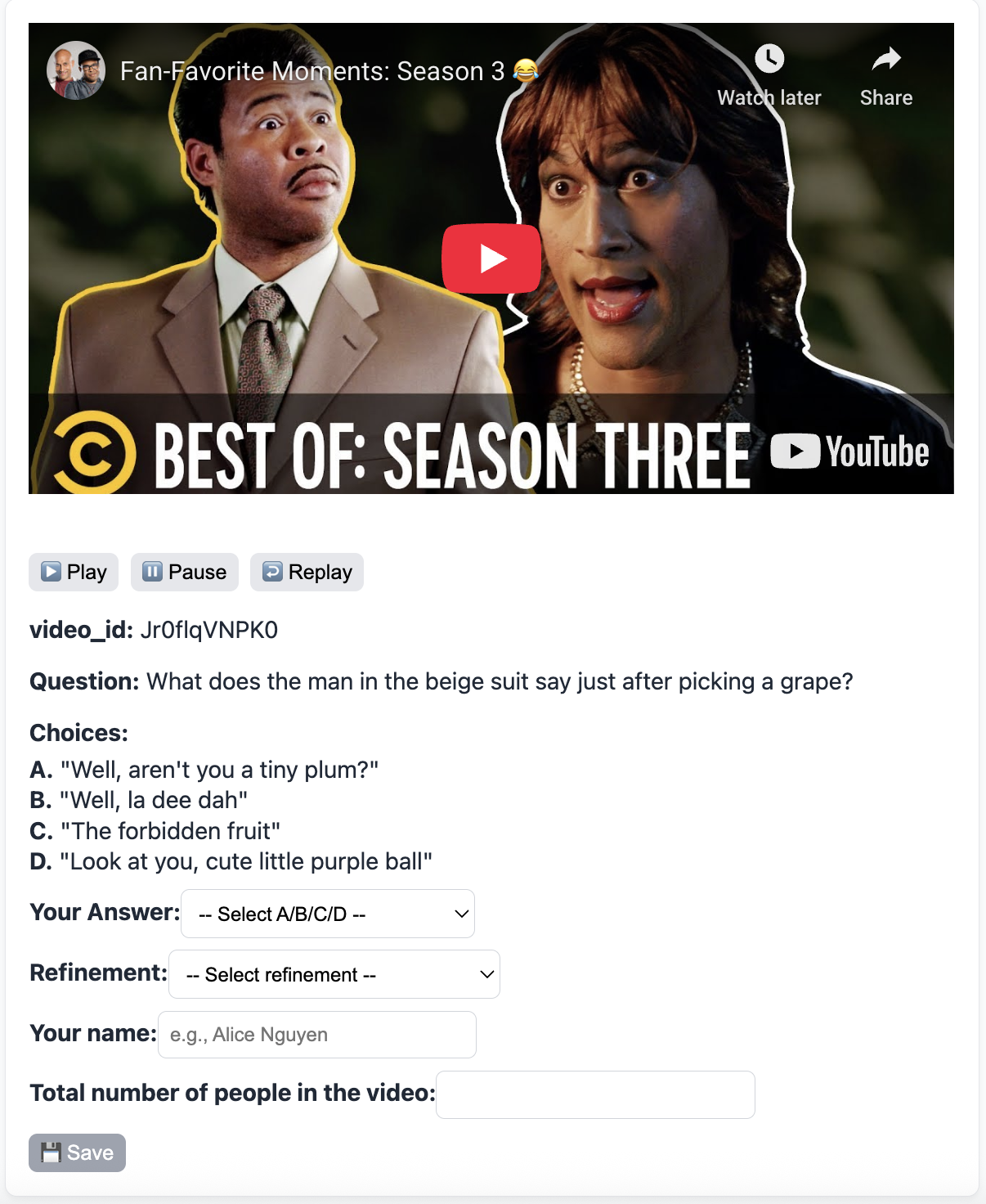}
    \caption{\textbf{Human evaluation interface.} Evaluators watch the video clip, then answer the corresponding multiple-choice question (A--D). 
    No transcript or subtitle is provided.
    The interface also includes an optional refinement tag and a control question asking for the total number of people visible in the video. 
    This setup ensures that human performance is independent of annotation and directly comparable to model outputs.}
    \label{fig:human_interface}
\end{figure}

\subsection{Human Performance}
\label{supp:human_perf}

To obtain a reliable human upper bound, we conduct a separate human evaluation that is fully decoupled from the dataset annotation workflow. 
None of the human evaluators participated in authoring or reviewing the dataset questions. 
The evaluation team consists of ten undergraduate and master's students, all with strong English proficiency. 
Each evaluator answered roughly 300 questions, providing broad, non-overlapping coverage of the test set. 
This design follows common practice in recent multimodal benchmarks and ensures that the reported human accuracy reflects genuine task difficulty rather than annotator familiarity.

\paragraph{Evaluation Interface and Protocol.}
Figure~\ref{fig:human_interface} shows the interface used for human evaluation. 
The interface presents the video clip, accompanied by playback controls (\emph{Play}, \emph{Pause}, \emph{Replay}), followed by the multiple-choice question exactly as posed to the model. 
Evaluators select their answer from four options (A--D) without access to transcripts, subtitles, or any auxiliary textual resources. 
To keep track on the number of speakers in each question, the interface also includes a control prompt asking evaluators to report the \emph{total number of people visible in the video}.

Each evaluator completes their assigned questions through the interface in random order. 
After submission, the interface immediately advances to the next item, preventing revisiting or revision. 
Evaluators do not have access to ground-truth labels or the responses of other participants. 
The interface additionally provides an optional refinement tag (e.g., \emph{trivial error}) that allows evaluators to flag questions with clear issues, such as mismatches between the question and answer options, or between the video and the question. These tags are used for data sanity checks and are not shown to models or used in computing human accuracy. In practice, no refinement tags were submitted during the human evaluation process.

This protocol ensures that human performance is (i) independent of the annotation process, (ii) based solely on audio--visual information, and (iii) directly comparable to model predictions under the same multiple-choice setting.

\section{Qualitative Analysis by Error Pattern}
\label{supp:error_example}

\begin{figure*}[t]
    \centering
    \includegraphics[width=\linewidth]{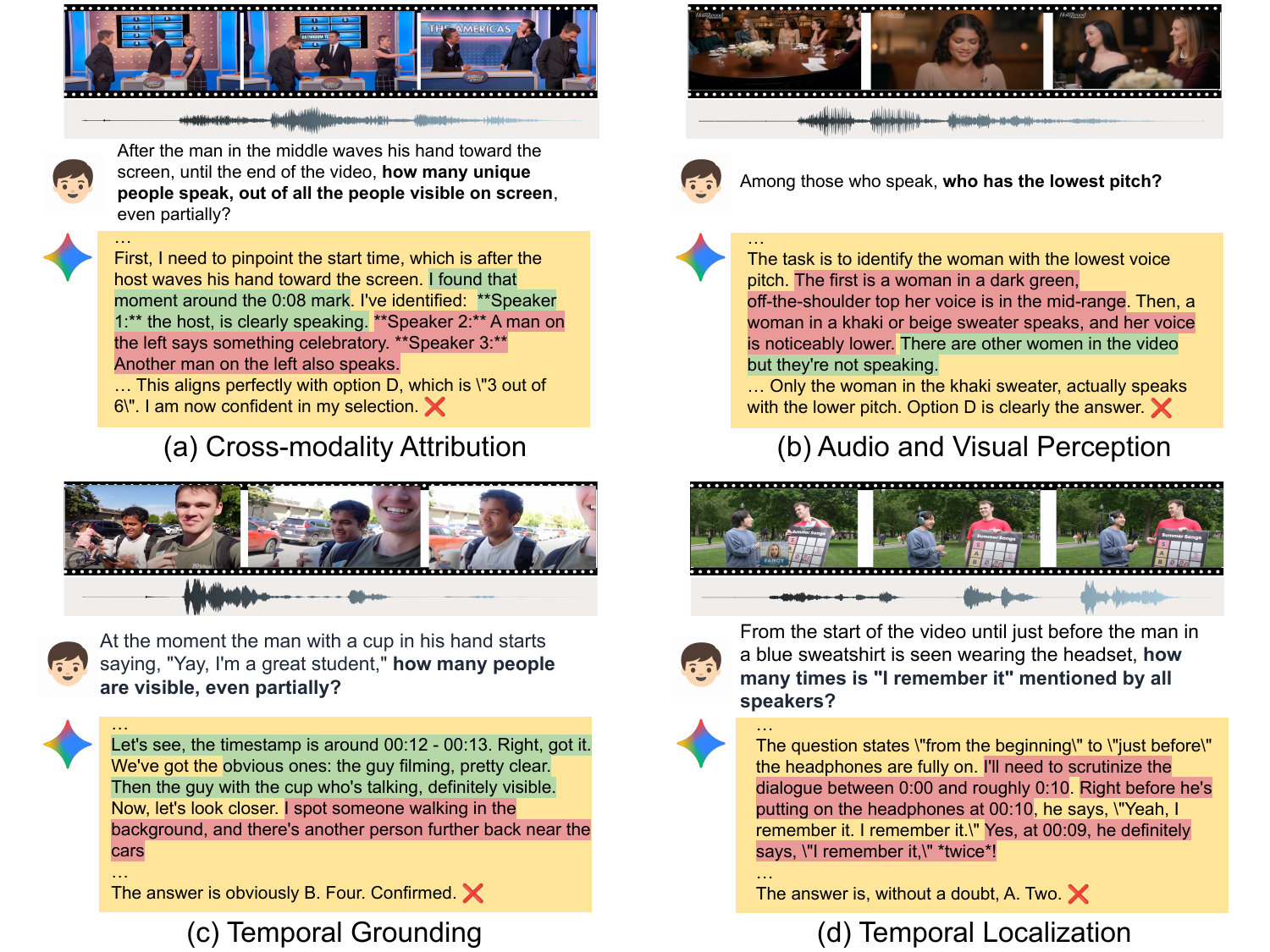}
    \caption{\textbf{Qualitative examples of Gemini 2.5 Pro} reasoning traces on \textbf{AV-SpeakerBench}. \textcolor{darkgreen}{Green} and \textcolor{darkred}{red} highlight colors indicate the model’s correct and incorrect reasoning, respectively. 
    The figure above contains representative failure cases spanning four key error patterns: (a) cross-modality attribution, (b) audio and visual perception, (c) temporal grounding, and (d) temporal localization. 
    Detailed analyses are provided in the subsection.}

    \label{fig:error_patterns}
\end{figure*}

Figure~\ref{fig:error_patterns} presents qualitative examples of Gemini~2.5~Pro’s reasoning traces on AV-SpeakerBench, organized by four representative error patterns. 
In all cases, the model has access to the full video and audio segment, yet its thinking reveals where the audiovisual reasoning breaks down.
\paragraph{(a) Cross-modality attribution.}
In Figure~\ref{fig:error_patterns}a, the question asks how many unique people speak after the host waves his hand toward the screen. 
The video shows that only the host and a woman respond verbally. 
However, the model attributes the utterance to the man standing on the left, and further hallucinates an additional off-screen male speaker. 
This error reflects a failure to align the acoustic evidence (voice timbre and timing) with the correct visual identity, leading to incorrect speaker attribution and over-counting of speakers.

\paragraph{(b) Audio and visual perception.}
The example in Figure~\ref{fig:error_patterns}b targets the person with the lowest voice pitch among several women. 
The model first misperceives the visual scene, describing a \emph{non-existent} ``woman in a dark green off-the-shoulder top'' instead of the actual woman in a blue dress. 
It then claims that the woman in a khaki sweater has the lowest pitch, even though her pitch is not the lowest in the clip. 
This case jointly exposes incorrect visual (wrong clothing description) and incorrect audio perception (wrong comparative pitch), showing that errors in one modality can reinforce errors in the other.

\paragraph{(c) Temporal grounding.}
The example in Figure~\ref{fig:error_patterns}c probes whether the model can restrict its reasoning to a short temporal window. 
The question asks how many people are visible, even partially, when the man with a cup in his hand speaks at the specific frame that the man starts saying the phrase.
The model correctly identifies this time span as the anchor, but then counts people who appear before (a person walking in the background) and after (a person farther back near the cars). 
The answer is therefore based on people outside the annotated window, illustrating a temporal grounding error: the anchor is found, but the visual evidence is aggregated over a broader interval than the question specifies.

\paragraph{(d) Temporal localization.}
In Figure~\ref{fig:error_patterns}d, the example focuses on the number of times ``I remember it'' is mentioned by all speakers from the beginning of the clip until a specific action: the man in the blue sweater putting on the headphones. 
In the video, the relevant action occurs at approximately 00:08, and the correct count is determined by utterances up to that moment. 
The model, however, localizes the anchor around 00:10 in its reasoning trace, shifting the cut-off point later in time.
This mis-localization changes which utterances are counted and leads to an incorrect answer, highlighting that even small temporal offsets can cause large reasoning errors in speech-counting tasks.

\section{Question Examples}
\label{supp:example}

To make the task definitions concrete, we visualize representative multiple–choice question–answer pairs from AV-SpeakerBench.  
Each example is shown as a panel containing (from top to bottom) a strip of video frames, the corresponding audio waveform, a natural–language question, and four answer options.  
Across all tasks, the model is evaluated by its accuracy in selecting the correct option, given the paired audio–visual clip.

\begin{figure*}[t]
    \centering
    \includegraphics[width=\linewidth]{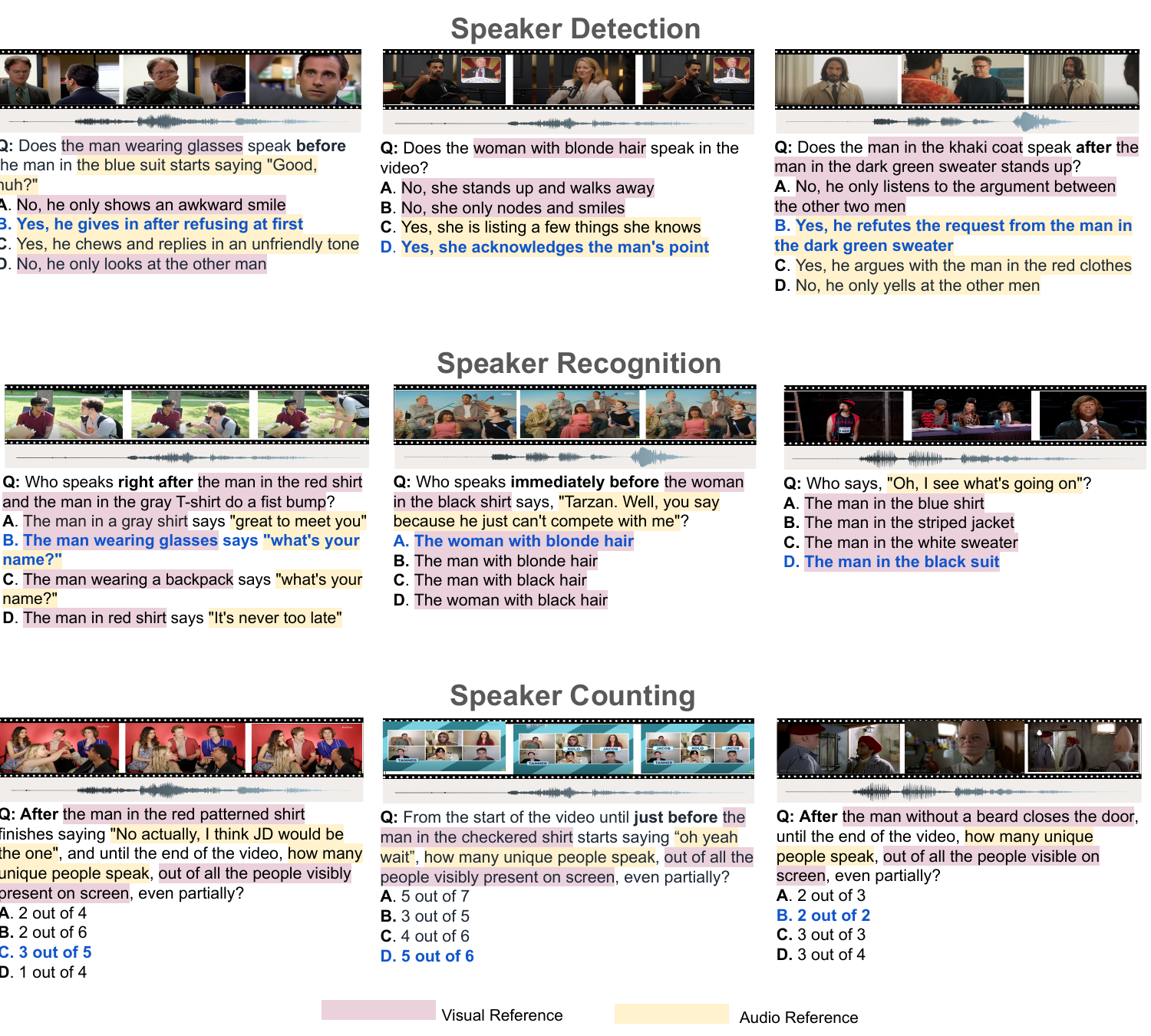}
    \caption{\textbf{Visualization of speaker tasks.} \textbf{Top:} Speaker Detection. \textbf{Middle:} Speaker Recognition. \textbf{Bottom:} Speaker Counting.}
    \label{fig:example_speaker_tasks}
\end{figure*}

Figure~\ref{fig:example_speaker_tasks} groups the \emph{speaker-centric} tasks: Speaker Detection, Speaker Recognition, and Speaker Counting.  
In these tasks, the evaluation focuses on whether the model can correctly associate speech segments with the corresponding visible speaker(s), distinguishing who is speaking, when they speak, and how many distinct speakers are active.

\begin{figure*}[t]
    \centering
    \includegraphics[width=\linewidth]{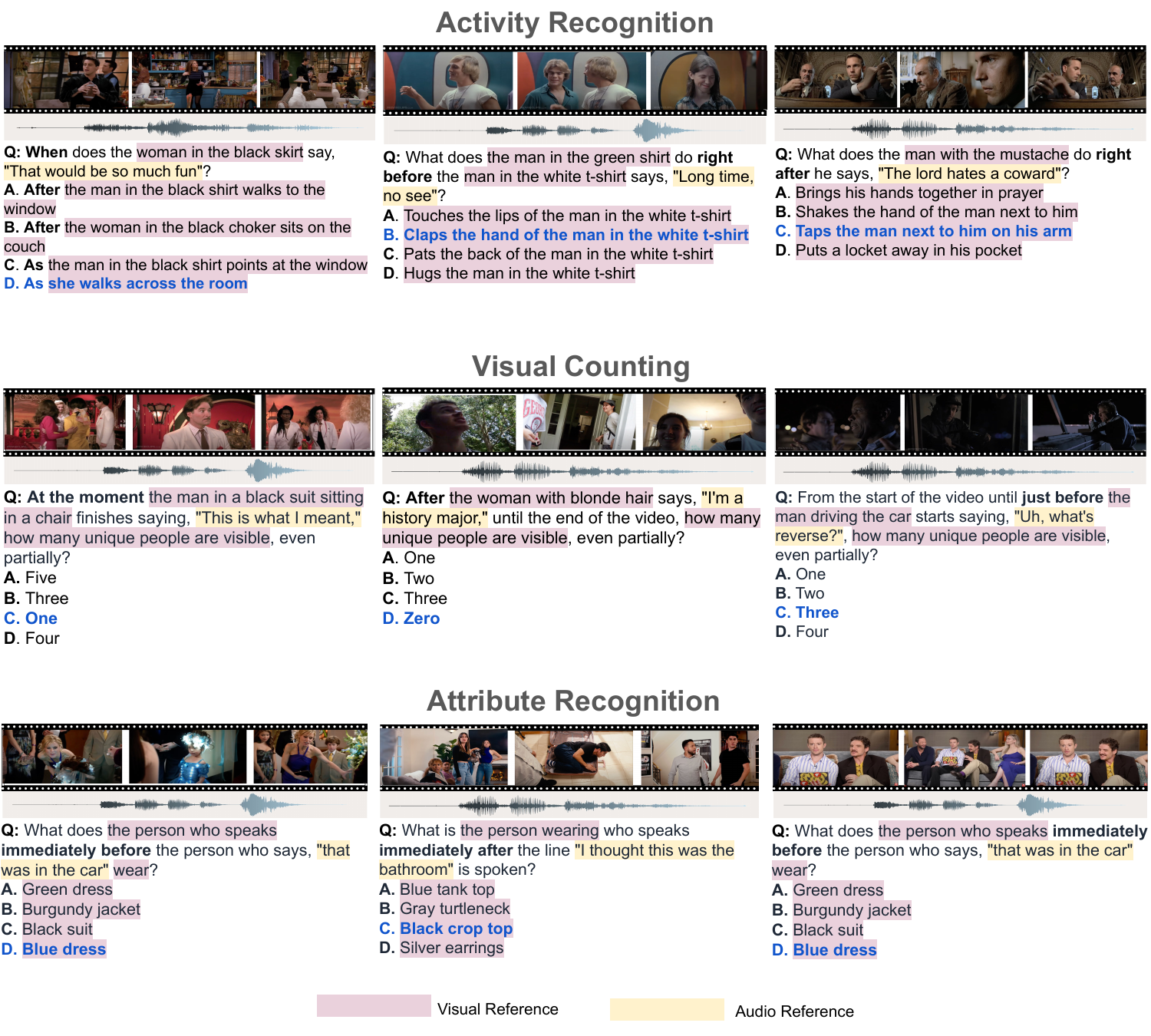}
    \caption{\textbf{Visualization of speaker-visual tasks.} \textbf{Top:} Activity Recognition. \textbf{Middle:} Visual Counting. \textbf{Bottom:} Attribute Recognition.}
    \label{fig:example_speaker_visual}
\end{figure*}

Figure~\ref{fig:example_speaker_visual} illustrates \emph{speaker–visual} tasks such as Activity Recognition and visually grounded counting.  
Here, the model must reason jointly about a speaking person and their on-screen actions or surrounding context (e.g., who performs an action while or around speaking, or how often a visually specified person speaks), coupling appearance, motion, and speech.

\begin{table*}[ht]
\centering
\setlength{\tabcolsep}{6pt}
\renewcommand{\arraystretch}{1.2}
\resizebox{\linewidth}{!}{
\begin{tabular}{lc|ccc|ccc|cccccc|c}
\toprule
\multicolumn{2}{c|}{} &
\multicolumn{3}{c|}{\textbf{Speaker-centric}} &
\multicolumn{3}{c|}{\textbf{Visual-centric}} &
\multicolumn{6}{c|}{\textbf{Audio-centric}} &  \multicolumn{1}{c}{}\\
\midrule
\multicolumn{2}{c|}{\textbf{Task Type}} &
Detection & Recognition & Counting &
Attribute Recognition & Activity Recognition & Counting &
Recognition & Duration & Pitch & Rate & Intensity & Counting & Overall\\
\midrule
Human Performance & -- & 96.02 & 93.13 & 94.28 & 93.14 & 93.20 & 94.15 & 96.52 & 90.68 & 93.20 & 91.39 & 94.17 & 93.40 & 93.74  \\
\midrule
Gemini 2.5 Pro Thinking~\cite{comanici2025gemini25pushingfrontier} & -- & 81.73 & 74.15 & 74.13 & 72.55 & 73.30 & 62.93 & 77.11 & 78.81 & 67.48 & 69.86 & 71.84 & 63.89 & 73.04\\
Gemini 3 Pro Thinking & -- & 85.95 & 79.86 & 73.13 & 80.39 & 79.13 & 71.71 & 87.06 & 75.85 & 70.87 & 78.47 & 75.24 & 70.14 & 77.62\\
\bottomrule
\end{tabular}
}
\caption{\textbf{Evaluation Results of Gemini 3 Pro (Thinking) on AV-SpeakerBench}. }
\label{tab:gemini3_result}
\end{table*}

\begin{table}[]
    \centering
    \begin{tabular}{l|ccc}
        \toprule
        \textbf{Model (Size)} & \textbf{8f} & \textbf{16f} & \textbf{1fps} \\
        \midrule
        Gemini 2.5P (-) & 70.4 & 72.0 & \textbf{73.0} \\ 
        Phi4-MM (5.6B)  & 38.5 & 39.2 & 39.5 \\
        VITA 1.5 (7B)   & 36.5 & 36.8 & 36.3 \\
        Qwen2.5-O (7B)  & 41.0 & 41.3 & 42.3 \\
        Qwen3-O (30B)   & 51.3 & 53.1 & 54.1 \\
        \bottomrule
    \end{tabular}
    
    \caption{\textbf{Performance on different sampling strategies.}}
    \label{tab:frame_sampling}
\end{table}

\begin{figure*}[t]
    \centering
    \includegraphics[width=\linewidth]{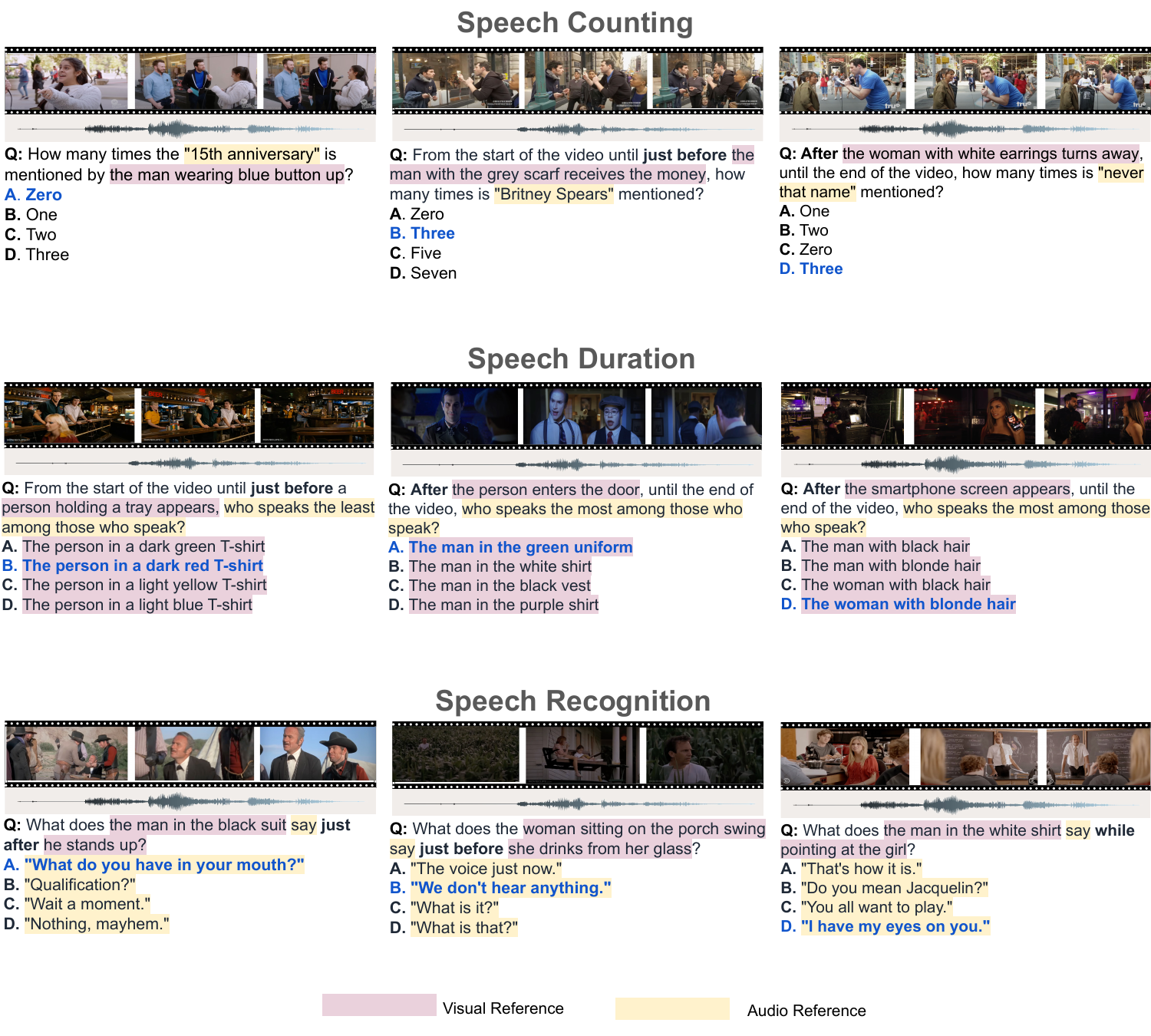}
    \caption{\textbf{Visualization of speech tasks.} \textbf{Top:} Speech Counting. \textbf{Middle:} Speech Duration. \textbf{Bottom:} Speech Recognition.}
    \label{fig:example_speech_tasks}
\end{figure*}

Figure~\ref{fig:example_speech_tasks} showcases \emph{speech reasoning} tasks: Speech Recognition, Speech Counting, and Speech Duration. The evaluation in these tasks requires recovering short spoken phrases, counting phrases, and compare the duration of speech.  

\begin{figure*}[t]
    \centering
    \includegraphics[width=\linewidth]{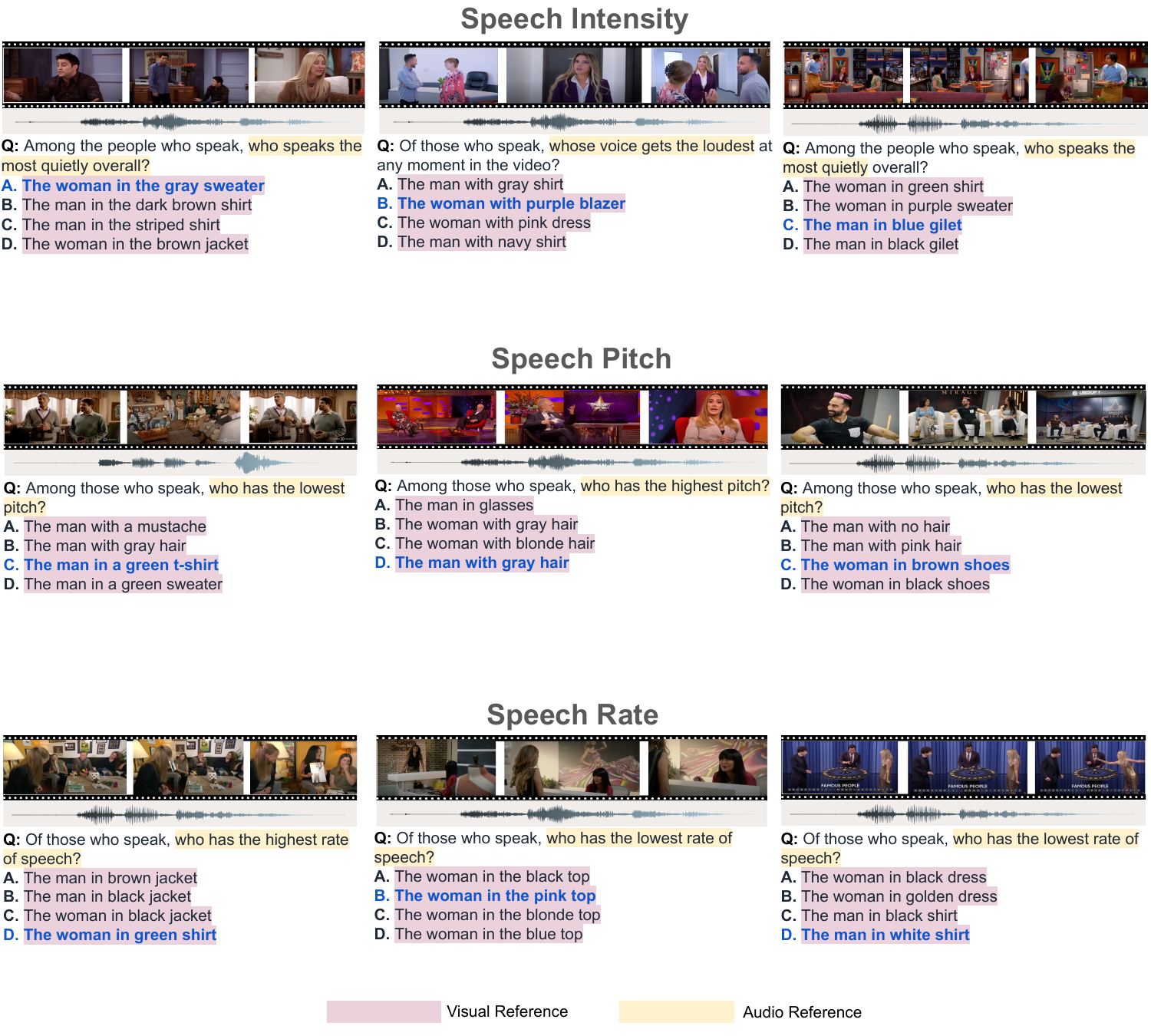}
    \caption{\textbf{Visualization of speech attribute tasks.} \textbf{Top:} Speech Intensity. \textbf{Middle:} Speech Pitch. \textbf{Bottom:} Speech Rate.}
    \label{fig:example_speech_attributes}
\end{figure*}

Finally, Figure~\ref{fig:example_speech_attributes} presents \emph{speech attribute} tasks, including  and comparative speech questions on rate, pitch, and intensity.  These examples emphasize fine-grained acoustic reasoning: the model must use the audio waveform to compare how different people speak (e.g., how fast, how high or loud).

For compactness, each figure arranges three examples per row and one row per task group, while maintaining a consistent interface across all panels.  Together, these four figures highlight the diversity of natural-language formulations and reasoning skills targeted by AV-SpeakerBench, spanning identity, actions, counting, appearance, speech, and attributes within a unified audio–visual QA framework.

\section{Performance of Gemini 3 Pro on AV-SpeakerBench}

We evaluate the recently released Gemini 3 Pro (Thinking)\footnote{Released after the main paper deadline.} and compare it with Gemini 2.5 Pro on AV-SpeakerBench.

As shown in Table~\ref{tab:gemini3_result}, Gemini~3~Pro attains an average accuracy of \textbf{77.62\%}, outperforming Gemini~2.5~Pro (73.04\%) by \textbf{+4.6} points. The improvements are broad and most pronounced in tasks requiring fine-grained audio–visual grounding, including speaker-centric understanding, visual reasoning, speech recognition, and prosody-related perception. However, several categories—most notably counting and speech-attribute reasoning—remain far from solved, and a substantial gap persists relative to human performance (93.74\%).

Overall, Gemini~3~Pro represents meaningful progress but still falls short of demonstrating robust and generalizable audiovisual reasoning.

\begin{figure}
    \centering
    \includegraphics[width=\linewidth]{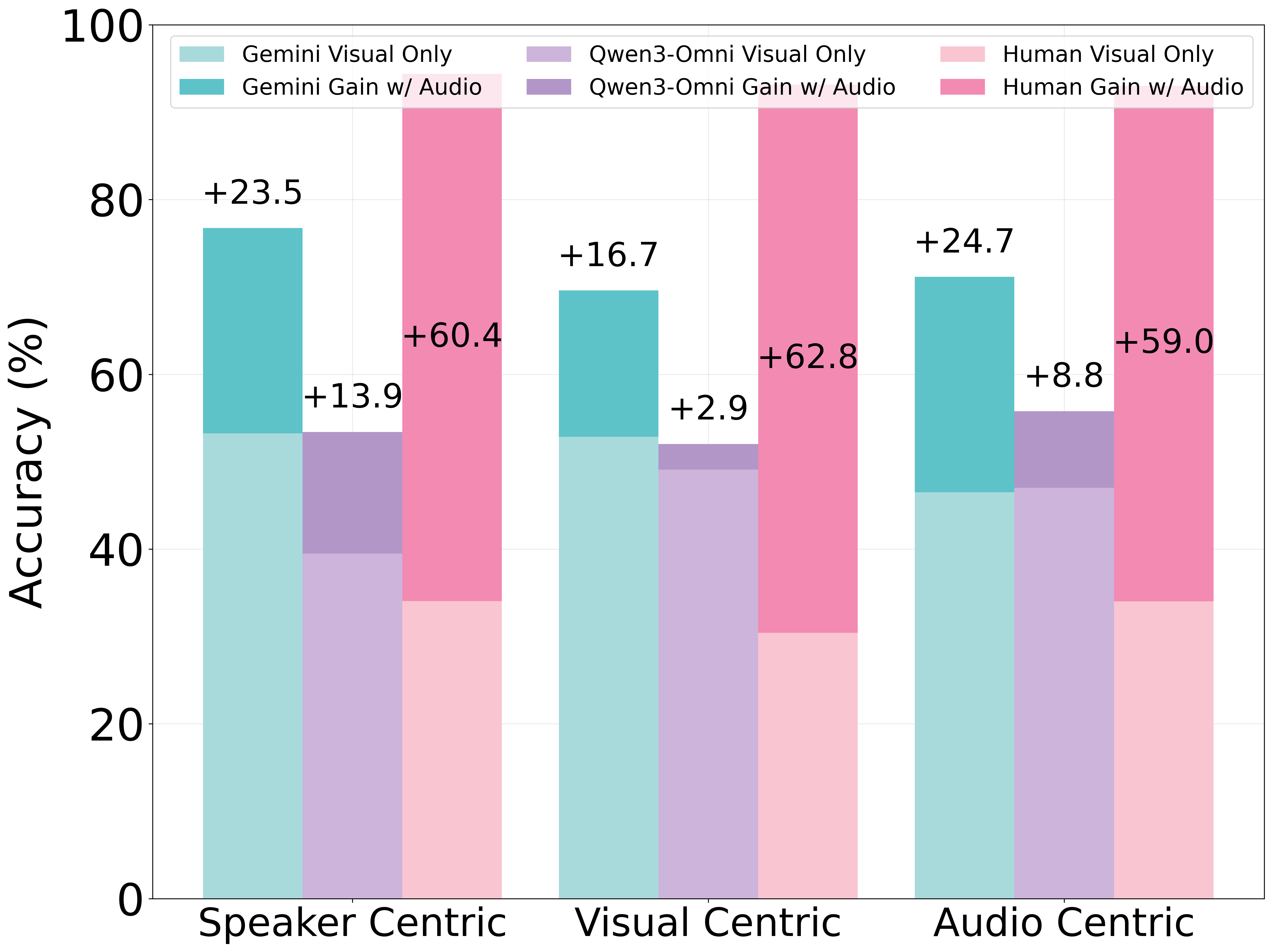}
    \caption{\textbf{Performance under visual-only (V) and audiovisual (A+V) settings.}}
    \label{fig:visual_only_rebuttal}
\end{figure}

\section{Evaluation beyond default frame sampling.}
\label{supp:frame_sampling}

We conducted additional experiments with models achieving non-trivial accuracy on AV-SpeakerBench under different frame sampling densities. Notably, the strongest models already adopt a similar default policy (1 fps with a frame cap). We report results with 8 and 16 frames since these 2 numbers are widely adopted in audiovisual large language model. Results show that these fixed number of frames consistently leads to slightly \emph{lower} accuracy, while preserving the same model ranking (Table~\ref{tab:frame_sampling}).

\section{Audio necessity vs.\ vision-only solutions.}
\label{supp:vision_only}
AV-SpeakerBench enforces fusion through question design, grounding speech to visible identities and temporal events. Human evaluation shows that the benchmark is largely \emph{not solvable} from vision alone, while audiovisual input yields near-ceiling performance (Figure~\ref{fig:visual_only_rebuttal}). In fact, we observe the same trend across all 12 tasks. Some items may still be inferred from natural visual conversational cues (e.g., expressions or gestures), but overall audiovisual fusion provides the most reliable and general solution. We further label questions that humans cannot answer under vision-only as an \emph{audio-required} subset; Gemini~2.5~Pro achieves 72.7\% on this subset versus 73.04\% overall. 

\vspace{-3pt}
\section{Ethical Statement}
\vspace{-3pt}

We define our dataset as a collection of publicly available YouTube videos paired with precise timestamps and human-written annotations. The benchmark is used solely for evaluation, not model training, and thus poses minimal risk of bias amplification or unintended memorization. The dataset will be released under the CC BY-NC-SA 4.0 license, which restricts use to non-commercial research and prohibits applications involving facial recognition, surveillance, or biometric identification. Individuals featured in referenced videos may request removal, and we will promptly withdraw the corresponding segments from future releases.

\clearpage

\onecolumn

\begin{longtable}{|p{0.18\linewidth} | p{0.78\linewidth}|}
\caption{Task-specific goals and requirements for all annotation tasks.}
\label{tab:task_requirements} \\
\toprule
\textbf{Task Name} & \textbf{Requirement} \\
\midrule
\endfirsthead

\toprule
\textbf{Task Name} & \textbf{Requirement} \\
\midrule
\endhead

\bottomrule
\endlastfoot

Speaker Detection & \textbf{Goal:} Test whether models can detect whether a person is currently speaking among multiple visible people.
\begin{itemize}[leftmargin=1.2em]
    \item Select a 5--30\,s clip with at least two visible individuals, where only one person is clearly speaking during a short moment.
    \item Choose a target person (e.g, the man in black suit), and a short utterance as an audio anchor or a short event as a visual anchor.
    \item Formulate a question such as: ``Does the [person] speak'' or ``Does the [person] speak before/after/when [utterance/event] occurs?'' 
    \item Provide four answer options, with 2 No and 2 Yes. 
    For each No option, put a visual reference (e.g, No, he just stares at the opposite person). 
    For each Yes option, put an audio reference (e.g, Yes, he talks about his plan).
    \item If there is not enough audio/visual reference, please create plausible distractors that fit the scene context.
\end{itemize}
\\
\midrule

Speaker Recognition &
\textbf{ Goal:} Evaluate whether models can link an utterance to the correct visible speaker.
\begin{itemize}[leftmargin=1.2em]
    \item Select a 5--30\,s clip with multiple visible speakers.
    \item Choose a short utterance as an audio anchor or a short event as a visual anchor.
    \item Formulate a question using: ``Who speaks right after/before/when [person] says [utterance]?'', ``Who responds to [person] saying [utterance]?'', or ``Who says [utterance]?'' 
    \item Provide four answer options referring to different individuals (e.g., ``the man in the black suit'', ``the woman in the red dress'').
    \item If fewer than three people appear, create plausible distractors that fit the scene.
\end{itemize}
\\
\midrule

Speaker Counting &
\textbf{ Goal:} Assess whether models can count how many distinct people speak in a segment.
\begin{itemize}[leftmargin=1.2em]
    \item Select a 10--30\,s clip with multiple visible people and several distinct spoken turns.
    \item Choose a short utterance as an audio anchor or a short event as a visual anchor.
    \item Formulate a question such as: ``From the start of the video until [utterance/event], how many people speak, out of the people in the video?'' or ``After [utterance/event] until the end of the video, how many people speak, out of the people in the video?''
    \item Provide four numerical answer options in the format “$n$ out of $m$” (e.g., “1 out of 2”, “2 out of 3”); the first number must be the true count of speakers in the specified interval, and the second number must be the number of visible people in that segment.
    \item Ensure that speakers are visually distinguishable and separated by clear turn-taking; create plausible nearby counts as distractors.
\end{itemize}
\\
\midrule

Attribute Recognition &
\textbf{ Goal:} Test whether models can connect a speaker’s appearance to their speech right before or after an anchor.
\begin{itemize}[leftmargin=1.2em]
    \item Select a 10--30\,s clip with multiple visible speakers.
    \item Choose a short utterance as an audio anchor.
    \item Formulate a question such as: ``What is the appearance of the person who says [immediately before/after/when] the person says [utterance]?'' or ``What does the person who says [utterance] wear?'' 
    \item Ask about attributes like clothes, clothing shape, clothing color, hair, hairstyle, or hair color.
    \item Provide four answer options describing different plausible appearances; exactly one must match the correct speaker.
    \item If fewer than four distinct appearances are visible, create plausible distractors consistent with the scene.
\end{itemize}
\\
\midrule

Activity Recognition &
\textbf{ Goal:} Measure whether models can associate a speaker with actions that happen before/after/while they speak.
\begin{itemize}[leftmargin=1.2em]
    \item Select a 10--30\,s clip with multiple visible speakers and at least one clear action (e.g., standing up, waving, pointing).
    \item Choose a short utterance as an audio anchor.
    \item Formulate a question such as: ``What does the person do [immediately before/after/when] the person says [utterance]?'' or ``When does the person say [utterance]?'' 
    \item For the first question type, encode the activity (e.g., ``stands up'', ``raises a hand'') into the four answer choices.
    \item For the second question type, encode [before/after/when] + [activity] inside the four answer choices.
    \item If fewer than three distinct activities appear, create plausible distractors consistent with the scene.
\end{itemize}
\\
\midrule

Visual Counting &
 \textbf{ Goal:} Assess whether models can count visible entities conditioned on an audio anchor.
\begin{itemize}[leftmargin=1.2em]
    \item Select a 10--30\,s clip with multiple visible people.
    \item Choose a short utterance as the reference time.
    \item Formulate a question such as: ``After/Before/When the [utterance] occurs, how many people are visible, even partly?''
    \item Provide four numerical answer options (e.g., 1, 2, 3, 4) or phrases that clearly encode the count.
    \item Ensure all counted entities are clearly visible in the frame; create plausible distractors using nearby counts.
\end{itemize}
\\
\midrule

Speech Recognition &
\begin{itemize}[leftmargin=1.2em]
    \item \textbf{ Goal:} Evaluate whether models can recognize the spoken content in the clip.
    \item Select a 10--30\,s clip with multiple visible speakers.
    \item Choose a short visual event as an anchor.
    \item Formulate a question such as: ``What does the person say [just before/after/when] the [event] occurs?'' 
    \item Provide four answer options corresponding to different possible transcripts; exactly one must match the spoken phrase (allow light paraphrasing only for the correct option).
    \item Ensure distractor phrases are grammatical and plausible in the context of the video.
\end{itemize}
\\
\midrule

Speech Duration &
\textbf{ Goal:} Check whether models can reason about relative speaking time.
\begin{itemize}[leftmargin=1.2em]
    \item Select a 5--20\,s clip where at least two people each speak for clearly different durations.
    \item Choose an event as a visual anchor.
    \item Formulate a question such as: ``Of those who speak, who speaks the most/least?'', ``From the start of the video, until [event], who speaks the most/least?'', or ``After [event], until the end of the video, who speaks the most/least?" 
    \item Provide four answer options referring to different individuals (e.g., ``the man in the black suit'', ``the woman in the red dress'').
    \item Ensure the duration difference is perceptible (at least 1--2\,s); construct plausible but incorrect visual options as distractors.
\end{itemize}
\\
\midrule

Speech Pitch &
\textbf{ Goal:} Test whether models can distinguish speakers by relative pitch.
\begin{itemize}[leftmargin=1.2em]
    \item Select a 5-10\,s clip where at least two speakers have clearly different pitch ranges.
    \item Formulate a question such as: ``Among those who speak, who has the highest/lowest pitch?'' 
    \item Provide four answer options referring to different individuals in the scene (e.g., ``the man in the black suit'', ``the woman in the red dress'').
    \item If fewer than three speakers are present, create plausible distractors based on other visible characters that could plausibly speak.
\end{itemize}
\\
\midrule

Speech Rate &
\textbf{ Goal:} Measure whether models can compare how fast different speakers talk.
\begin{itemize}[leftmargin=1.2em]
    \item Select a 5--10\,s clip with at least two speakers whose speaking rates differ noticeably.
    \item Formulate a question such as: ``Among those who speak, who has the highest/lowest rate of speech?'' 
    \item Provide four answer options referring to different speakers (e.g., by clothing or position).
    \item Ensure the difference in speaking rate is clear (e.g., more syllables in similar time), and create plausible distractors when needed.
\end{itemize}
\\
\midrule

Speech Intensity &
\textbf{ Goal:} Assess whether models can reason about speech loudness.
\begin{itemize}[leftmargin=1.2em]
    \item Select a 5--10\,s clip where at least two speakers talk with clearly different loudness levels (e.g., one shouting, one speaking softly).
    \item Formulate a question such as: ``Among those who speak, who has the highest/lowest voice?''
    \item Provide four answer options referring to different individuals (e.g., ``the man in the black suit'', ``the woman in the red dress'').
    \item Avoid clips where microphone distance alone explains loudness, unless this is visually clear; create plausible distractors as needed.
\end{itemize}
\\
\midrule

Speech Counting &
\textbf{ Goal:} Evaluate whether models can count number of phrases or keywrods.
\begin{itemize}[leftmargin=1.2em]
    \item Select a 5--30\,s clip with multiple short utterances or turns of speech.
    \item Choose an event as a visual anchor.
    \item Formulate a question such as: ``From the start of the video, until just before [event], how many times [phrase] is mentioned by [person/everyone]?" or ``After [event], until the end of the video, how many times [phrase] is mentioned by [person/everyone]?" 
    \item Provide four numerical answer options; exactly one must match the true count.
    \item Ensure utterances are separated enough to be countable (clear pauses or turn-taking), and use nearby counts (e.g., $n-1$, $n$, $n+1$, $n+2$) as distractors.
\end{itemize}
\\

\end{longtable}


\end{document}